\documentclass[10pt,twocolumn,letterpaper]{article}

\usepackage{cvpr}              % To produce the CAMERA-READY version
\usepackage{amsmath,amsfonts,bm}

\def\eqref#1{equation~\ref{#1}}
\def\1{\bm{1}}

\DeclareMathAlphabet{\mathsfit}{\encodingdefault}{\sfdefault}{m}{sl}
\SetMathAlphabet{\mathsfit}{bold}{\encodingdefault}{\sfdefault}{bx}{n}

\newcommand{\ANURAG}[1]{\textcolor{blue}{\bf \small [Anurag: #1]}}
\newcommand{\MICHAEL}[1]{\textcolor{red}{\bf \small [Michael: #1]}}

\newcommand{\TODO}[1]{\textcolor{blue}{\bf \small [To do: #1]}}

\usepackage{url}

\usepackage{booktabs}       % professional-quality tables
\usepackage{amsfonts}       % blackboard math symbols
\usepackage{nicefrac}       % compact symbols for 1/2, etc.
\usepackage{microtype}      % microtypography
\usepackage{xcolor}         % colors
\usepackage{algorithm2e}
\usepackage{graphicx,wrapfig}

\usepackage{caption} 
\usepackage{subcaption}

\usepackage{balance}
\usepackage[pagebackref,breaklinks,colorlinks]{hyperref}

\def\cvprPaperID{9232} % *** Enter the CVPR Paper ID here
\def\confName{CVPR}
\def\confYear{2023}

\begin{document}

\title{Token Turing Machines}

\author{Michael S. Ryoo, Keerthana Gopalakrishnan, Kumara Kahatapitiya, Ted Xiao,\\ Kanishka Rao, Austin Stone, Yao Lu, Julian Ibarz, Anurag Arnab\\
Google Research \\
%\texttt{\{mryoo,\}@google.com}
\texttt{mryoo@google.com}
}
\maketitle

\begin{abstract}
%We propose Token Turing Machines for real-world sequential decision making. Token Turing Machine is a modernization of the previous Neural Turing Machine, which is a differentiable sequential model with external memory interactions. Motivated by Turing machines, Neural Turing Machine proposed differentiable operations to maintain and utilize external memory (a matrix in this case) using a neural network controller, sequentially. In Token Turing Machines, the external memory is maintained as a set of tokens, and the model learns memory read and write operations as token learning processes. Transformers are used as the processing unit. We demonstrate the power of TTM on two complicated visual learning tasks requiring sequential reasoning: (1) temporal activity localization from videos (i.e., activity detection) and (2) vision-based robot action policy learning (i.e., visuo-motor policy learning)).

We propose Token Turing Machines (TTM), a sequential, autoregressive Transformer model with memory for real-world sequential visual understanding.
Our model is inspired by the seminal Neural Turing Machine, and has an external memory consisting of a set of tokens which summarise the previous history (i.e., frames).
This memory is efficiently addressed, read and written using a Transformer as the processing unit/controller at each step.
The model's memory module ensures that a new observation will only be processed with the contents of the memory (and not the entire history), meaning that it can efficiently process long sequences with a bounded computational cost at each step.
We show that TTM outperforms other alternatives, such as other Transformer models designed for long sequences and recurrent neural networks, on two real-world sequential visual understanding tasks: online temporal activity detection from videos and vision-based robot action policy learning.

Code is publicly available at:
\href{https://github.com/google-research/scenic/tree/main/scenic/projects/token_turing}{https://github.com/google-research/scenic/tree/main/scenic/projects/token\_turing}.
%\url{https://github.com/google-research/scenic/tree/main/scenic/projects/token_turing}

\end{abstract}

\section{Introduction}

Processing long, sequential visual inputs in a causal manner is a problem central to numerous applications in robotics and vision. For instance, human activity recognition models for monitoring patients and elders are required to make real-time inference on ongoing activities from streaming videos. As the observations grow continuously, these models require an efficient way of summarizing and maintaining information in their past image frames with limited compute. Similarly, robots learning their action policies from training videos, need to abstract history of past observations and leverage it when making its action decisions in real-time. 
%robot action policy learning requires a robot to abstract history of past observations and leverage this when making decisions in real-time.
This is even more important if the robot is required to learn complicated tasks with longer temporal horizons.
%Similar capabilities are required by computer vision systems analysing human activities in streaming videos.

A traditional way of handling online observations of variable sequence lengths is to use recurrent neural networks (RNNs), which are sequential, auto-regressive models~\cite{hochreiter1997long,chung2014empirical, donahue2015long}.
As Transformers~\cite{vaswani2017attention} have become the de facto model architecture for a range of perception tasks, several works have proposed variants which can handle longer sequence lengths~\cite{dai2019transformer, tay2020efficient, wu2022memorizing}.
However, in streaming, or sequential inference problems, efficient attention operations for handling longer sequence lengths themselves are often not sufficient since we do not want to run our entire transformer model for each time step when a new observation (e.g., a new frame) is provided.
This necessitates developing models with explicit memories, enabling a model to fuse relevant past history with current observation to make a prediction at current time step.
Another desideratum for such models, to scale to long sequence lengths, is that the computational cost at each time step should be constant, regardless of the length of the previous history.

In this paper, we propose Token Turing Machines (TTMs), a sequential, auto-regressive model with external memory and constant computational time complexity at each step. 
%\ANURAG{If we highlight ``constant computational cost at each time step'', we need experiments and other things to reinforce this point}.\MICHAEL{Problem is that any model could become a constant cost if we brute-forcely cut the window?}\ANURAG{constant cost whilst having full receptive field? That is not so easy then. And MemViT does not have this for example}\MICHAEL{But our window is just 6 or 8 due to the TPU memory...}\ANURAG{Hehe. What we can really claim is that the FLOPs / RAM usage are the same for step 1 as it is for step 8 ... RNNs have this. Some other transformer models do ont.} \MICHAEL{I agree, but I am not sure what the best way to show this would be.}
Our model is inspired by Neural Turing Machines~\cite{graves2014neural} (NTM), an influential paper that was among the first to propose an explicit memory and differentiable addressing mechanism.
The original NTM was notorious for being a complex architecture that was difficult to train, and it has therefore been largely forgotten as other modelling advances have been made in the field.
However, we show how we can formulate an external memory as well as a processing unit that reads and writes to this memory using Transformers (plus other operations which are common in modern Transformer architectures).
Another key component of TTMs is a token summarization module, which provides an inductive bias that intuitively encourages the memory to specialise to different parts of its history during the reading and writing operations.
Moreover, this design choice ensures that the computational cost of our network is constant irrespective of the sequence length, enabling scalable, real-time, online inference.

In contrast to the original NTM, our Transformer-based modernisation is simple to implement and train.
We demonstrate its capabilities by achieving substantial improvements over strong baselines in two diverse and challenging tasks: (1) online temporal action detection (i.e., localisation) from videos and (2) vision-based robot action policy learning.

\section{Related Work}

Our TTM model is related to prior work on designing Transformers to process long sequence lengths and temporal context (e.g., videos), and also models to store and retrieve relevant information from internal/external memory.

\vspace{-5pt}
\paragraph{Transformers for sequences.}

Pairwise self-attention mechanism proposed in Transformer \cite{vaswani2017attention} has been very successful in many vision tasks, including the understanding of image sequences. Extending ViT \cite{dosovitskiy2020}, ViViT \cite{arnab2021vivit} and TimeSformer \cite{bertasius2021space} represented video data, a series of space-time tokens. Transformers have since become state-of-the-art in video modeling, being able to handle multiple modalities like audio/text \cite{nagrani2021attention, akbari2021vatt} and scaling efficiently \cite{fan2021multiscale, liu2022video}.

One of the major challenges in using Transformers for sequential data is the well-known quadratic computation cost of self-attention. That is, as the number of frames in a video sequence increase, the computation grows quadratically which often soon becomes intractable.
%The quadratic cost of self-attention is well known.
There is a wide body of work on reducing this to enable transformers to handle longer sequence lengths, as summarized in surveys such as~\cite{tay2020efficient,tay2020long}.
Common themes include local- or sparse-attention~\cite{zaheer2020big, liu2021swin, child2019generating}, pooling or reducing the number of tokens within the network~\cite{ryoo2021tokenlearner, jaegle2021perceiver, rae2019compressive} and approximations of the attention matrix~\cite{choromanski2020rethinking, wang2020linformer, peng2021random}.

However, in the sequential inference problems considered in this paper, efficient operations for handing longer sequence lengths are often not sufficient themselves, as we do not want to perform redundant operations at every new time step, when new input tokens are given.

\vspace{-5pt}
\paragraph{Transformers with memories.}
One manner of reducing redundancy over time-steps is to leverage models with memory.
There are a number of works using Transformers to retrieve relevant information from external memories/knowledge bases~\cite{khandelwal2019generalization,borgeaud2022improving,guu2020retrieval,wu2022memorizing} or historical observations \cite{burtsev2020memory, le2019learning, rae2019compressive}.
In comparison, the memory of our model is also based on historical observations of the model, which informs the current and future predictions; we learn to maintain/read/write to the memory.
Another method for reusing computation from previous time steps is to perform causal attention. 
%, where the model only attends to tokens from previous time-steps.
In this case, the previous activations of the model can be cached, as done in the original implementation of the Transformer~\cite{vaswani2017attention}.
However, with this approach, the computation cost at each step still linearly increases over time as the sequence length of previous tokens increase.
Transformer-XL~\cite{dai2019transformer} builds upon this idea, and uses relative positional embeddings to make better use of previous history tokens.
MemViT~\cite{wu2022memvit} also uses token activations from previous time-steps to increase the contextual information provided at the current time step.
However, once again, the computational cost at each step still increases over time.
\cite{goyal2022coordination,didolkar2022temporal} introduced an approach to improve pairwise operations in Transformers by interacting with (external) shared workspace, which also could be viewed as memory read/write.

\vspace{-5pt}
\paragraph{Sequential models.}

The classical solution for dealing with long and variable sequence lengths are recurrent neural networks, which share the same parameters across multiple time-steps to be able to generalise to varying sequence lengths.
LSTMs~\cite{hochreiter1997long} and GRUs~\cite{chung2014empirical} are the most well-known form of these networks, as they were formulated to handle the ``vanishing and exploding gradient'' problem~\cite{hochreiter2001gradient}.
Video representations using them also have been common, traditionally \cite{donahue2015long}.
Transformers have been adapted to recurrent networks as well, with models such as Block-Recurrent Transformers~\cite{hutchins2022block}, which is an RNN with a transformer operating on a sequence (or block) of tokens, instead of traditional RNNs which have a single previouss state. \cite{kaiser2017learning} adds an external memory module to LSTMs/GRUs. 

\vspace{-5pt}
\paragraph{Neural Turing Machines.}
Our model, however, is based on another formulation of RNNs, the Neural Turing Machine (NTM)~\cite{graves2014neural}.
This model architecture is based on the von Neumann computer architecture, and consists of a controller and external memory which is read and written to using explicit addressing operations in a differentiable manner. Its memory access mechanism was further extended in the subsequent work, DMC \cite{graves2016hybrid}. The original NTM was a complex model that was notorious for being difficult to train.
Our formulation can be thought of as a modernisation of this architecture using Transformer-based operations as primitives.
Our model is simple and easy to train, and we have also applied it to complex problems in computer vision (and visual robot learning) that the original NTM was never demonstrated for.

\begin{figure*}
    \centering
    \includegraphics[width=0.75\linewidth]{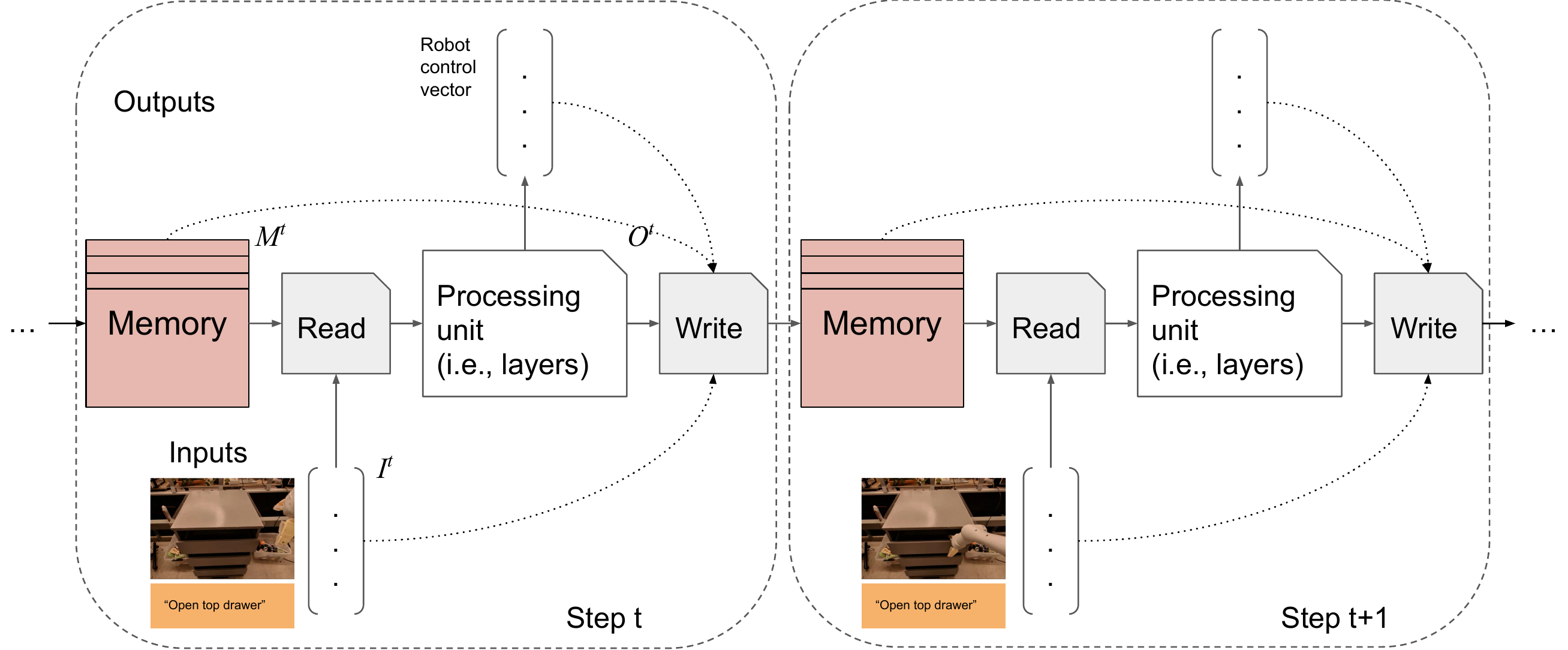}
    \caption{TTM overview with robot examples. Each dotted rectangle shows TTM at each step.
    % \ANURAG{We can use an expanded version of this as the ``teaser'' figure?}
    % \MICHAEL{What do we exactly mean with an expanded version? More timesteps? More details? Fancier illustrations? I also moved this to be at the beginning of the section rather than at the end.}
    % \ANURAG{Fancier :-) Also, somehow show the task? And that the output is used at each step?}
    % \ANURAG{Also, could include some of the notation that we use in the method section here.}
    % \MICHAEL{Maybe we can provide Charades frames before the inputs and provide activity annotations as the outputs?}
    % \ANURAG{Yeah, or the robotics one. Depends which one our experiments focus on}
    % \ANURAG{We also need to differentiate this with our RNNs / memory networks.}
    }
    \label{fig:overview}
\end{figure*}
%0.8

\section{Token Turing Machines}

%\MICHAEL{This is an alternative version of the technical section.}

%\REPHRASE{Token Turing Machines can be seen as a different type of recurrent neural network},

Token Turing Machines are new sequential auto-regressive models, with core components being the (external) memory and the processing unit, as shown in Fig.~\ref{fig:overview}.
The memory at time step $t$, $M^t \in \mathbb{R}^{m \times d}$, consists of a set of $m$ tokens of dimensionality $d$.
In Token Turing Machines (TTMs), the interface between the processing unit and memory are done purely in terms of ``read'' and ``write'' operations.
The result of memory ``read'' is fed to the processing unit. The output from the processing unit is ``written'' to the memory.

As illustrated in Fig.~\ref{fig:overview}, the input at time step $I^t \in \mathbb{R}^{n \times d}$, is merged with the memory $M^t$ to retrieve relevant tokens from both, which are then \emph{processed} further to produce $O^t \in \mathbb{R}^{r \times d}$.
The outputs of this step, along with the previous inputs and current memory are then used to write to the memory, $M^{t+ 1}$, which will be used at the next time step.
As many sequential decision making tasks require predictions at each time step, we also include a linear output head at each step.

\subsection{Memory Interface}

% A core principle of our approach is that 

There are several principles which motivate the design of our memory interface:
Intuitively, we do not wish to read from (or write to) all memory tokens at each time step.
This is because although the memory should contain (summarised) information from the entire past history, only some of this information may be relevant for the processing at the current stage.
Therefore, we consider selective reading of a smaller subset of tokens to be a good inductive bias that will encourage the model to make use of a memory that stores relevant information over varying time scales Such ``selective reading'' is in contrast to most previous recurrent models (e.g., RNNs) directly digesting history vectors.

Moreover, there may be redundancies in the input stream, $I^t$, due to the information that we already have in our memory, $M^t$, and due to the data itself (for example, videos contain redundant frames; not all parts of an image are relevant to the task at hand).
Therefore, a mechanism to summarise tokens, both from the memory and the input stream, is a core component of our approach.
We discuss this summarisation procedure next in Sec.~\ref{sec:token_summarise} before describing reading (Sec.~\ref{sec:method_read}), processing (Sec.~\ref{sec:method_process}) and writing (Sec.~\ref{sec:method_write}).

\subsubsection{Token Summarisation}
\label{sec:token_summarise}

There are multiple methods of summarising a sequence of $p$ tokens with dimensionality $d$, $V \in \mathbb{R}^{p \times d}$, to $Z \in \mathbb{R}^{k \times d}$ where $k \ll p$. 
%\MICHAEL{Do we mean $k \ll j$? By the way, if we can find any other symbol than $j$ we can use here, it will be great. We already used m, n, ...}
Examples include \cite{ryoo2021tokenlearner, jaegle2021perceiver, yin2022vit, rao2021dynamicvit, fayyaz2022adaptive, cordonnier2021differentiable} which have been proposed in the context of more efficient transformer backbones for processing higher-resolution images.
We adopt a similar approach as a core component of our reading and writing mechanisms.
Our method is based on \cite{ryoo2021tokenlearner, jaegle2021perceiver} motivated by the fact that these approaches are simple, fully-differentiable and have achieved strong results in a number of domains.

Concretely, we summarise a set of tokens $V$ by computing an importance weight vector, $w_i \in \mathbb{R}^{p}$ which we use to compute a weighted summation over the $p$ tokens. 
Note that we have $w_i$ for each output token, $i \in \{1, \ldots, k \}$, and it is computed with a learnable function taking the input $V$ itself, $\alpha_i(V)$.
%Here, we compute each importance weight either using a MLP function or using a learned query vector, $q_i$, computed as:
Here, such each importance weighting function is modeled either using a MLP or using a learned query vector, $q_i$, computed as:
\begin{align}
\label{eq:alpha}
    w_i &= \alpha_i(V) = \text{softmax}(\text{MLP}(V)), ~~\text{or}\\
    w_i &= \alpha_i(V) = \text{softmax}(q_i V^\top / \sqrt{d}).
\end{align}
These weights are then used to perform a weighted summation of the inputs:
\begin{equation}
    z_i = s_i(V) = w_i \cdot V = \alpha_i(V) \cdot V,
\end{equation}
where each token $z_i$ summarises all the tokens from the complete set $V$, based on the dynamic weighting $w_i = \alpha_i(V)$.
As we learn to summarize $p$ tokens into $k$ tokens, it computes a matrix $W = [w_1, \cdots, w_k]$ of importance weights in practice.
%this means that we learn a matrix $Q \in \mathbb{R}^{k \times d}$ to obtain a matrix $W$ of importance weights in practice.
%Therefore, this summarisation procedure is equivalent to cross-attention~\cite{vaswani2017attention} with $k$ learned query vectors.

Overall, we denote this summarisation function as $S_k: \mathbb{R}^{p \times d} \to \mathbb{R}^{k \times d}$, which we use for both memory read and write.

\subsubsection{Reading from Memory}
\label{sec:method_read}
\begin{figure}
    \centering
    \includegraphics[width=0.85\linewidth]{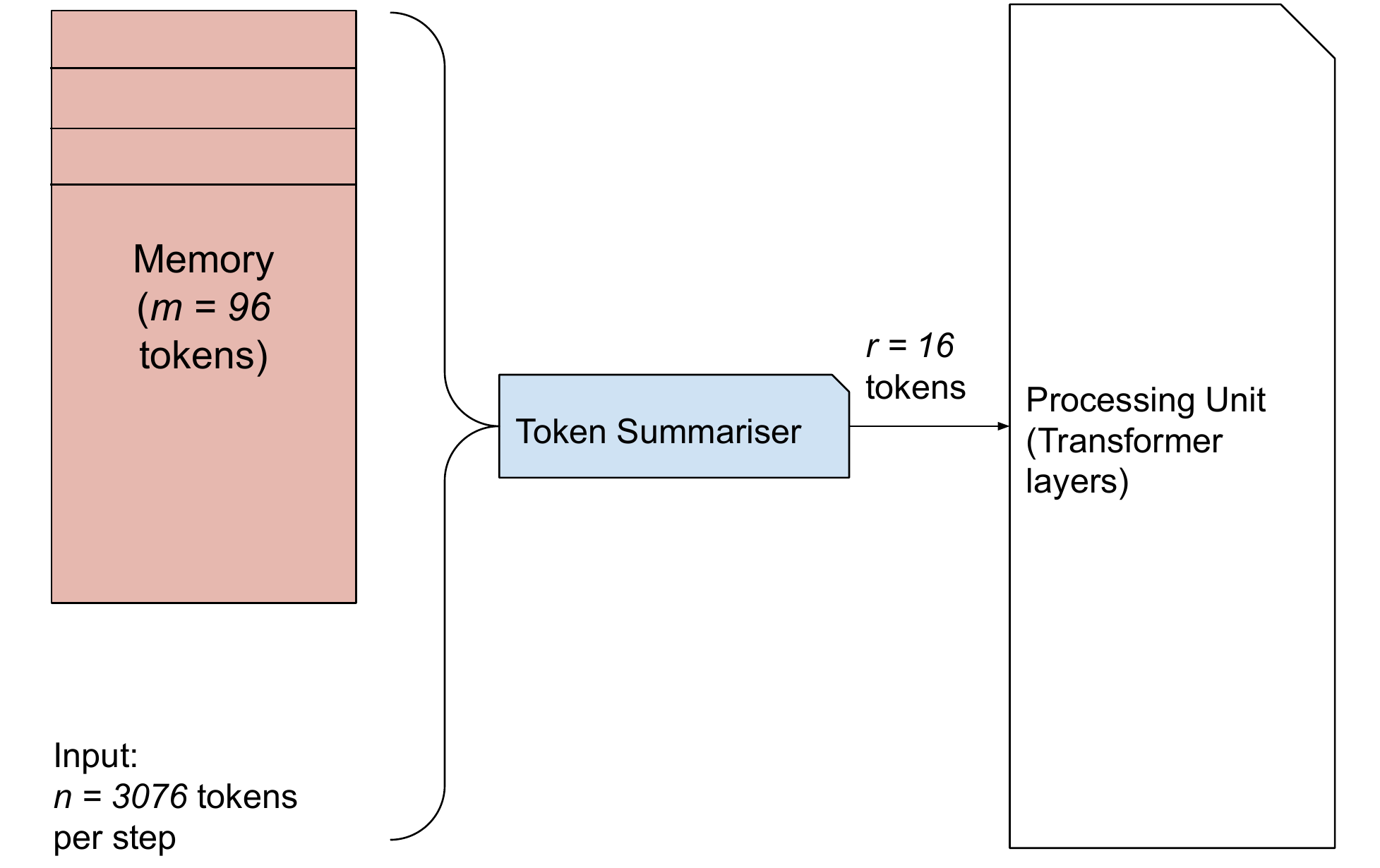}
    \caption{TTM Read. Note how it greatly reduces the computation of the subsequent processing module by summarising the input sequence as well.}
    \label{fig:read}
\end{figure}
% 0.9

In contrast to Neural Turing Machines~\cite{graves2014neural}, where inputs and memory are separately processed and merged later, we take a unified memory-input reading strategy.
This is motivated by the fact that some of the inputs, $I^t$, are redundant given the information that we already have in memory.

As illustrated in Fig.~\ref{fig:read}, we concatenate the tokens in memory, $M^t$ composed of $m$ tokens, with the input stream, $I^t$ composed of $n$ tokens, and summarise these tokens into a smaller subset of $r$ tokens.
Our read operator is thus defined as
\begin{equation}
    Z^t = \text{Read}(M^t, I^t) = S_r([M^t || X^t]),
\end{equation}
where $[M^t || X^t]$ denotes the concatenation of these two matrices. This essentially is a function of $\mathbb{R}^{(n+m) \times d} \to \mathbb{R}^{r \times d}$.
Thus, the read operator filters the information in the memory and input which should be passed to the subsequent processing unit.
Note that by reducing the number of tokens passed to the processing module, we also substantially reduce the computational cost of this stage.

\paragraph{Memory addressing by location using positional embedding} 

In principle, the token summarisation module described above enables content-based addressing of the memory. This was referred as ``read by content'' in the Neural Turing Machines. In order to also make the model take advantage of locations of the tokens within the memory (and also to distinguish tokens from memory vs. tokens from inputs), we add a learnable positional embedding \cite{dosovitskiy2020} before each read module. This approach, fusing position information into the tokens, has an effect of read/write by location 
(+ content) without modifying the overall process.

\subsubsection{Processing Unit}
\label{sec:method_process}

Our processing unit is a generic function, $O^t = \text{Process}(Z^t)$, that operates on the $r$ tokens obtained from the read operation, $Z^t$.
The processing function generates a set of $r$ output tokens, $O^t$, which are used in the subsequent write operation.
Moreover, for tasks which require a prediction at each time-step, we add a linear output-head $Y^t = \text{Output}(O^t) = W_o O^t$ to the output tokens.

In our experiments, we use a standard Transformer~\cite{vaswani2017attention} and MLPMixer~\cite{mlpmixer2021} as our processing unit, although other architectures are possible too.

\iffalse
    \ANURAG{This is copied from the previous version.}
    
    The proposed Token Turing Machine is generic and is able to cope with various types of processing units as long as they operate on top of tokens. We use a Transformer [] or MLPMixer [] architectures as our processing unit. Our processing unit corresponds to the `controller' in the Neural Turing Machine architecture. It is responsible for generating a set of output tokens, $O_t$, given the results of the memory+input read, $Z_t$. Specifically, $O_t = F(Z_t)$ where $F$ is the processing function. Any differentiable neural network function could be used as $F$, as long as it takes a set of tokens (or a matrix formed by concatenating token vectors) as an input and generates outputs also in the form of a set of tokens. These output tokens are also used to decide tokens to be written to the memory, as discussed above.
\fi

\subsubsection{Writing to Memory}
\label{sec:method_write}
\begin{figure}
    \centering
    \includegraphics[width=0.99\linewidth]{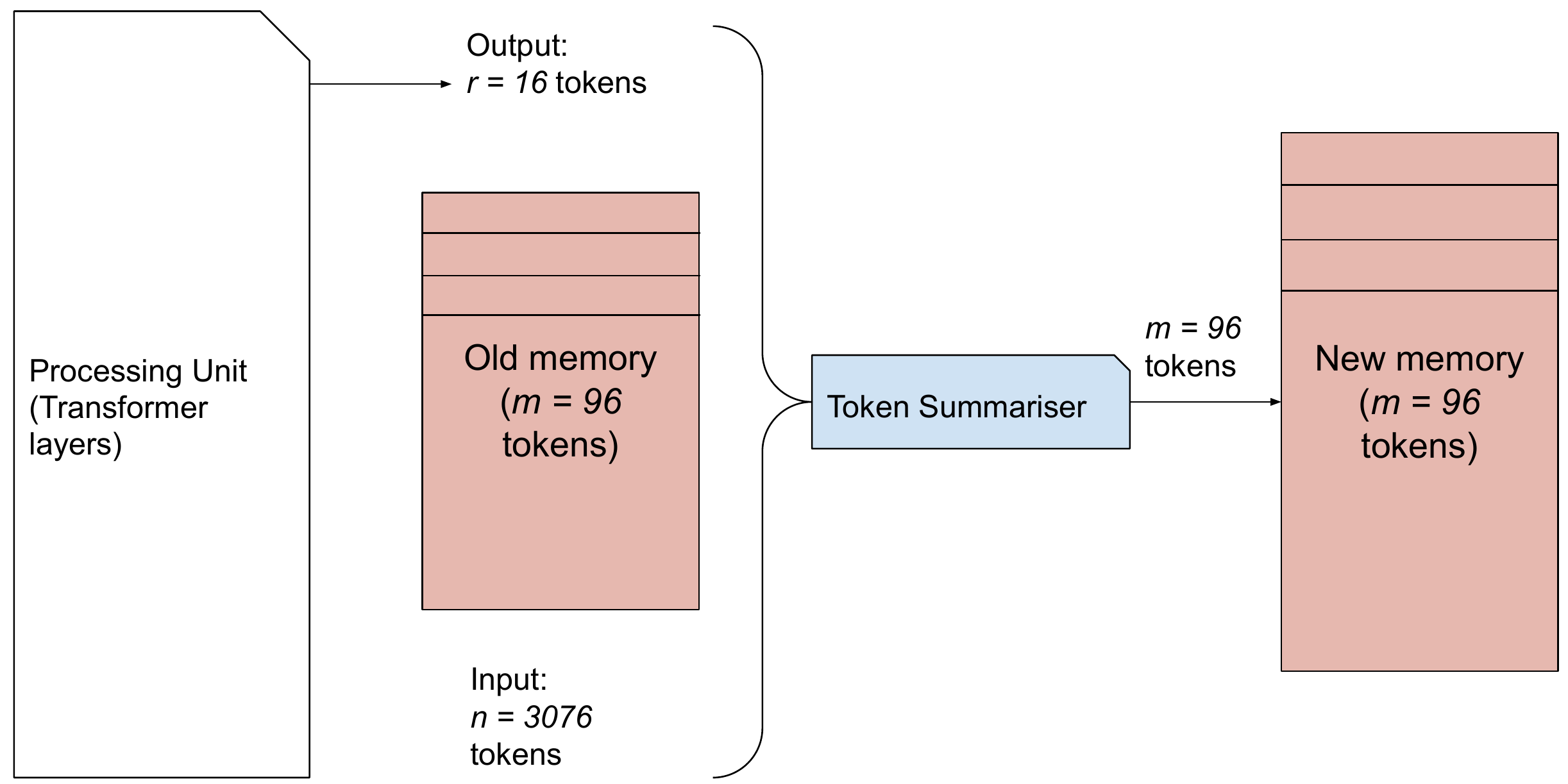}
    \caption{TTM Write, formulated as the token summarisation.}
    \label{fig:write}
\end{figure}
%0.99

We also formulate our write operation as a token summarisation process, which we observed to be both simple and effective.

As illustrated in Fig.~\ref{fig:write}, our write mechanism preserves tokens in the memory, $M^t$, by learning to re-select them.
And it adds new tokens to the memory by selecting them from either the output of the processing module, $O^t$, or from the inputs $I^t$. Therefore, we formulate our write operations as selecting $n$ tokens (i.e., the size of the memory) from the concatenation of current memory, input and output tokens, as denoted by
\begin{equation}
    M^{t + 1} = \text{Write}(M^t, O^t, I^t) = S_n([M^t || O^t || I^t]).
\end{equation}
Therefore, the tokens in memory will be erased if they are not re-selected. Similar to the read operation, positional embedding is used to distinguish tokens from memory, input, and output. This essentially is a function of $\mathbb{R}^{(n+m+r) \times d} \to \mathbb{R}^{m \times d}$.

\iffalse
    We formulate the memory write operation as a memory re-reading process. \TODO{Why? Because its simple and differentiable?}
    
    Our write mechanism preserves tokens in the memory by learning to re-select them, and it adds new tokens to the memory by learning to select tokens from elsewhere. That is, by repeating the token selection process $n$ times from a pool of (1) the tokens in memory $M_t$, (2) the input tokens $I_t$, and (3) the output tokens from the processing unit $O_t$, TTM constructs a new memory $M_{t+1}$ of size $n$.
    \begin{equation}
    O_t = \left\{A_i(M_t \cup I_t \cup O_t)\right\}_{i=1}^{n}.
    \end{equation}
    %($n$ is the size of the memory)
    The tokens in the memory will be erased if they are not (re-)selected. Fig.~\ref{fig:write} illustrate our memory write process.
\fi

% \subsection{Training}

% How do we train the model?
% ie how many steps do we backprop through?
% Can we do teacher forcing?
% Can we train on one sequence length and test on a larger one?

% \subsection{Implementation details}
% \ANURAG{Maybe part of the ``training'' section?}

\subsection{Discussion}

% \ANURAG{Maybe this is a good place to describe the differences, after having described our model in detail.}
% \MICHAEL{Definitely. We already have some text in go/token-machines.}
% \MICHAEL{Anurag, could you check what the best way to summarize what's in the google doc would be?}
% \ANURAG{Sure}

Our proposed Token Turing Machine can be viewed as a modernisation of Neural Turing Machines (NTM)~\cite{graves2014neural} by using Transformer-based models for the processing unit and its interfaces with the (external) memory.

Our reading and writing mechanisms differ to NTM in that we use token summarisation (Sec.~\ref{sec:token_summarise}) as the core component that unifies the two operations.
NTM, on the other hand, uses a complex range of ``content'' and ``location'' addressing strategies to produce the indices in the memory to read and write to, and learns a combination of matrix additions and deletions for the memory modifications.

And whilst we use a Transformer (or a Mixer) for the processing unit, NTM used either fully-connected feedforward or LSTM networks for its processing unit (or controller). The architectural choices of NTM meant that it was difficult to train in practice.
On the other hand, we have not witnessed training instabilities with our TTM model in the experiments that we present next.

Both TTMs and NTMs in general could be viewed as a new form of recurrent neural networks. The operations in TTMs can be summarised, in the recurrent network form as:
\begin{align}
    Z^t &= \text{Read}(I^t, M^t) \\
    O^t &= \text{Process}(Z^t) \\
    M^{t + 1} &= \text{Write}(M^t, O^t, I^t) \\
    Y^t &= \text{Output}(O^t)
\end{align}
where the functions Read(), Process(), Write(), and Output() are what we discussed in this section.
\section{Experiments}

\subsection{Video Activity Detection}

Activity Detection in videos focus on making fine-grained action predictions per every time step. In general, datasets for detection~\cite{caba2015activitynet, sigurdsson2016hollywood, yeung2018every, gu2018ava},  contain long-range videos with multiple overlapping activities, capturing an expressive temporal context. Hence, detection is more challenging compared to classification, which only makes a prediction once per video. Also, detection models often need to look at more frames to generate good representations in temporal context, which can be computationally expensive, particularly in Transformer-based architectures due to their quadratic cost.

We focus on making online inferences, generating activity predictions for each incoming frame. 
That is, the decision is made without accessing frames in future steps.
%considering only its previous temporal context. 
%At each timestep, models-- even Transformers with temporal causal masks-- need to process all previous frames together with the incoming frame, cost of which grows at least linearly with time.~\ANURAG{This is not correct. For causal transformers, you can cache the previous outputs, and then you don't need to calculate the self-attention for that. This is very common in NLP since it was there since \cite{vaswani2017attention}, and there is a flag in Flax that enables this actually for the Self-Attention layer.}\MICHAEL{I agree. Probably Kumara wrote?}

%With explicit memory, the proposed TTM can avoid this by having a constant complexity thanks to its read/ write mechanism. In fact, it only needs to process memory and the current frame to make predictions, which is much more efficient.

\subsubsection{Dataset and Settings}

Charades dataset~\cite{sigurdsson2016hollywood} contains $\sim$9.8k videos of 157 daily household activities, separated into $\sim$7.9k training and $\sim$1.8k validation clips. Each video may include multiple overlapping activities (w/ an average of 6.8 activity instances per video), annotated with frame-level labels. The average length of a video is 30 seconds. This is a challenging setting, especially for temporal activity detection, as a model needs to predict multiple potential activity classes per each frame, considering the interactions between different activities and longer temporal context.

For our evaluation, we use the standard `charades\_v1\_localize' setting, where we uniformly sample 25 frames from each video in the validation set and compute mean average precision (mAP).

We also use AVA v2.2 \cite{gu2018ava} as the secondary dataset to confirm the effectiveness of the TTM's sequential modeling, specifically in spatio-temporal activity detection. AVA is a dataset composed of bounding box annotations of 80 atomic visual actions in 430 15-minute movie clips. We follow its standard setting, while using the Kinetics-400 for the backbone pretraining.

\subsubsection{Baselines and Implementation}

We use ViViT~\cite{arnab2021vivit} as our backbone.
As was done in its original work, we made ViViT represent 32 frame segments. Given a continuous sequence of frames, ViViT converts it into a sequence of representations where each element is from a 32-frame segment. We use ViViT-B with its original settings: the input frame resolution is 224-by-224, and the video patch size is $16 \times 16 \times 2$ (i.e., an image patch of $16 \times 16$ over $2$ frames). It generates 14-by-14-by-16 (i.e., 3136) tokens, and this becomes our `step input' for TTMs: $n = 3136$. Alternatively, we do spatial average pooling per frame, getting 16 tokens per step: $n = 16$

%Token Turing Machines takes such element representation as its step input.

We tested various memory sizes ($m$), and we use $m = 96$ as the default setting. The number of reads is $r = 16$ (or $32$). The processing unit in TTMs is implemented to have a small overhead. In the case of using Transformers or MLPMixers, we used the hidden size of 512 and a total of four blocks. The training of the models was done by providing video segments of 6 steps (i.e., $32 \times 6$ frames) at a time. We include the detailed training settings in Appendix.

\begin{table}
  \centering
  \resizebox{1.\linewidth}{!}{
  \begin{tabular}{l|c|c|cc}
        \toprule
        Method & Setting &modality & mAP\\
        \midrule
        
        I3D + super-events~\cite{piergiovanni2018learning} & offline & RGB + Flow & 19.41 \\
        I3D + super-events + TGM~\cite{piergiovanni2019temporal} & offline & RGB + Flow & 22.30 \\
        I3D + STGCN~\cite{ghosh2020stacked} & offline & RGB + Flow & 19.09 \\
        I3D + biGRU + VS-ST-MPNN~\cite{mavroudi2020representation} & offline & RGB + Object & 23.7 \\
        Coarse-Fine (w/ X3D)~\cite{kahatapitiya2021coarse} & offline & RGB & 25.1 \\
        I3D + CTRN~\cite{dai2021ctrn} & offline & RGB & 25.3 \\
        I3D + MS-TCT~\cite{dai2022ms} & offline & RGB & 25.4 \\
        I3D + PDAN~\cite{dai2021pdan} & offline & RGB + Flow & 26.5 \\
        I3D + CTRN~\cite{dai2021ctrn} & offline & RGB + Flow & 27.8 \\
        \midrule
        I3D~\cite{carreira2017quo} & online & RGB + Flow & 17.22 \\
        X3D~\cite{feichtenhofer2020x3d} & online & RGB & 18.87 \\
        ViViT-B~\cite{arnab2021vivit} & online & RGB & 23.18 \\
        ViViT-B + TTM (ours) & online & RGB & 26.34 \\
        ViViT-L~\cite{arnab2021vivit} & online & RGB & 26.01 \\
        ViViT-L + TTM (ours) & online & RGB & 28.79 \\
        \bottomrule
  \end{tabular}
  }
  \caption{Comparison with the state-of-the-art methods on Charades temporal activity detection.}
  \label{table:charades_sota}
\end{table}

\subsubsection{Temporal Activity Detection Results}

In Table \ref{table:charades_sota}, we compare TTMs with prior state-of-the-art in temporal activity detection on Charades. The previous work we compare against include multiple backbone architectures (e.g., I3D~\cite{carreira2017quo}, X3D~\cite{feichtenhofer2020x3d}) as well as different techniques for long-term temporal modeling on top of the backbones (e.g., super-events~\cite{piergiovanni2018learning}, TGM~\cite{piergiovanni2019temporal}, Coarse-Fine~\cite{kahatapitiya2021coarse, kahatapitiya2021self}, and MS-TCT~\cite{dai2022ms}). 

Importantly, we grouped the approaches based on whether they support online inference or not. While most of the backbone models enable online inference by focusing on recent frames at hand, some approaches require a longer temporal window including future frames (e.g., the global snapshot of the entire video) to make a prediction for a given frame.

In addition, we compared TTMs with different sequential/temporal modeling architectures, applied on top of the same backbone model we use (i.e., ViViT). These include temporal Transformers and temporal MLPMixers~\cite{mlpmixer2021}, as well as more traditional sequential models like a LSTM. In addition, we implement a recurrent version of Transformers, which takes state tokens and input tokens to predict the next state tokens and output tokens. For temporal Transformers and MLPMixers, a temporal window of 6 steps is used. The models are over a fixed window, capturing $6 \times 32 = 192$ frames. The raw output from ViViT has $14 \times 14 \times 16$ tokens per step, giving us a total of $3136 \times 6 = 18816$ input tokens. We used aggressive (spatial) pooling and (temporal) striding to make their computational cost as low as TTMs, and FLOPS become comparable.

%we pooled one token per frame, generating 96 tokens, as it becomes computational intractable without such pooling (~19K tokens without pooling). We also made these models to process 16+16 tokens, sampling 16 tokens from prior ViViT step and 16 tokens from the current step.

Table \ref{table:temporal-comparison} shows the results. The FLOPS described are per-step inference time, excluding the backbone computation. We are able to confirm that TTM, due to its external memory interactions, enables much more efficient online inference compared to other types of sequential/temporal models. When TTM-Transformer and causal/recurrent Transformers are using the same number of input tokens (i.e., $n = 16$), TTM spends around 1/2 FLOPS compared to the causal/recurrent Transformers (TTM 0.228 vs. causal/recurrent Transformers 0.523/0.410 GFLOPS), while TTM still outperforms them (TTM 26.24 vs. causal/recurrent Transformers
25.85/25.97 mAP). In addition, while other temporal models have difficulty scaling (i.e., with a large number of tokens) due to overfitting, TTM handles more tokens much more reliably.

%TTMs achieve reasonable accuracies with much smaller computation (e.g., GFLOPS of causal Transformer vs. TTM is 0.523 vs. 0.089 with similar mAP). When a sufficient amount of computation is provided, i.e., when many input tokens are used, TTMs perform better while using less compute. The other temporal models like causal Transformers and MLPMixers have difficulty scaling due to the overfitting.

\begin{table}
  \centering
  \resizebox{1.\linewidth}{!}{
  \begin{tabular}{l|ccc}
        \toprule
        Method & mAP & GFLOPS \\
        \midrule
        ViViT only  & 23.18 & - \\
        \midrule
        \multicolumn{3}{l}{\textit{Alternative temporal models}} \\
        Temporal MLPMixer (tokens=96) & 24.41 & 0.382 \\
        Causal Transformer (tokens=96) & 25.85 & 0.523 \\
        Temporal Transformer (tokens=96) & 25.61 & 1.269 \\
        \midrule
        Temporal MLPMixer (tokens=3360) & 24.26 & 13.317 \\
        Causal Transformer (tokens=3360) & 25.88 & 29.695 \\
        Temporal Transformer (tokens=3360) & 25.53 & 112.836 \\
        \midrule
        \multicolumn{3}{l}{\textit{Alternative recurrent networks}} \\
        LSTM & 23.96 & 0.107 \\
        Recurrent Transformer (tokens=16+16) & 25.97 & 0.410 \\
        Recurrent Transformer (tokens=3136+16) & 25.97 & 17.10 \\
        \midrule
        \multicolumn{3}{l}{\textit{Token Turing Machines}} \\
        TTM-Mixer ($n = 16$) & 25.83 & 0.089 \\
        TTM-Transformer ($n = 16$) & 26.24 & 0.228 \\
        TTM-Mixer ($n = 3136$) & 26.14 & 0.704 \\
        TTM-Transformer ($n = 3136$) & 26.34 & 0.842 \\
        \bottomrule
  \end{tabular}
  }
  \caption{
    TTM vs. different sequence modeling methods. ViViT-B was used as the backbone. TTM-Transformer means we use Transformer as the processing unit, and TTM-Mixer means we use MLPMixer as the processing unit. FLOP measure is for computation in addition to the backbone.
    % \ANURAG{Do you only want to report the FLOPs in addition to the backbone?} \MICHAEL{Maybe?} 
    % \ANURAG{LSTM is worse than no temporal model?}
    % \MICHAEL{Yeah, overfitting? wrong hyper params?}
    % \ANURAG{What is the $r$ for Temporal Transformer and Mixer? There is no memory there?}
    % \MICHAEL{In this case, it's the total number of input tokens to Transformers/Mixers.}
    % \ANURAG{Are the input tokens reduced using TokenLearner, or just something like average pooling?}
    % \MICHAEL{spatial pooling + temporal sampling}
    %\ANURAG{The $n = 3136$ and $n = 96$ for TTM is confusing.}\MICHAEL{Let's talk a bit in a chat.}
  }
  \label{table:temporal-comparison}
\end{table}

\subsubsection{Ablations}

Here, we conduct a number of ablations to investigate different components of Token Turing Machines. We use ViViT-B as the backbone. Unless specified, the models use Transformer processing units by default, and MLP-based token summarisations. The number of input tokens per step is $n = 3136$.

%We also ran ablations comparing Token Turing Machines with...

%LSTM? ViViT + token learning? 

%\textbf{Size of the memory?} How many tokens do we need? Does performance plateau after a number of tokens? How does this depend on the task?

\vspace{-5pt}
\paragraph{Processing units:}
Table~\ref{table:unit} compares TTMs with different processing units. The default processing unit, i.e., Transformer, is compared against MLPMixer and a simple MLP. We observe that MLPMixer-based TTM provides a good speed-accuracy trade-off.

%\textbf{Controller architecture} Transformer with token learner, standard transformer, vanilla RNN, so on. How do each of this perform?

\begin{table}
\begin{minipage}{0.49\textwidth}
  \centering
  \begin{tabular}{l|cc}
        \toprule
        Architecture & mAP & GFLOPS \\
        \midrule
        MLP & 23.34 & 0.689 \\
        MLPMixer & 26.14 & 0.704 \\
        Transformer & 26.34 & 0.842 \\
        \bottomrule
  \end{tabular}
  \caption{Using different processing unit architectures in TTMs.}
  \vspace{3mm}
  \label{table:unit}
\end{minipage}
\begin{minipage}{0.49\textwidth}
  \centering
  \begin{tabular}{l|cc}
        \toprule
        Method & mAP & GFLOPS \\
        \midrule
        Pooling & 25.75 & 0.206 \\
        MLP & 26.34 & 0.842 \\
        Latent query & 26.75 & 8.537 \\
        \bottomrule
  \end{tabular}
  \caption{Different Token Summarisation}
  \label{table:tl}
\end{minipage}
\end{table}

\vspace{-5pt}
\paragraph{Token summarisation:}
Table~\ref{table:tl} compares different token summarisation methods used within TTMs. Essentially, we are comparing different form of the $\alpha_i$ function in Equation~\ref{eq:alpha}, which influences both memory read and write in TTMs.
We compare the MLP-based $\alpha$, the latent query-based $\alpha$, and a simple pooling-based summarisation (i.e., no learning) method.

%Another possibilty is to replace the TokenLearner with simpler pooling strategies, ie average pool to $k$ tokens or something like this.

\begin{table}
  \centering
  \resizebox{1.\linewidth}{!}{
  \begin{tabular}{l|cc}
        \toprule
        Method & mAP & GFLOPS \\
        \midrule
        % Pooling + sampling & 21.63 & 0.103 \\
        Concatenate (Memorizing Transformer-style) & 20.97 & 0.920 \\
        Erase and Add (NTM-style write) & 25.86 & 0.423 \\
        \midrule
        TTM without memory & 22.65 & 0.842 \\
        TTM & 26.34 & 0.842 \\
        \bottomrule
  \end{tabular}
  }
  \caption{TTM vs. different history/memory update. They all use Transformer processing units, and MLP-based token summarisations. The number of input tokens per step, $n = 3176$.}
  \label{table:memory}
\end{table}

%\textbf{Sequence lengths} How long are the sequence lengths during training. Do we have training instabilities? Can we test with longer sequence lengths than during training?

%\textbf{Computational benefits} As we are summarising the input sequence as well, we are also saving a lot of compute. Some experiments to show / quantify this.

%Btw, another baseline is to only perform TokenLearner on the memory, but not the input sequence when reading. So this let's us use memory, but we don't get any computation advantage from it.t

\vspace{-5pt}
\paragraph{Different memory read/write:}

We compare memory read/write mechanisms of TTMs with their alternatives, motivated by prior work including \cite{graves2014neural,wu2022memorizing}. Specifically, we implemented the memory write of concatenating every observed input tokens. 
%, with and without spatial pooling per frame.
We also implemented the memory write mechanism designed in \cite{graves2014neural}: write by erase and addition. Finally, a memory-free version of TTM was implemented to confirm the importance of the memory. This was done by zeroing out the memory of the TTM after each step, making it spend exactly the same amount of computation. Table~\ref{table:memory} shows the results.

% \begin{table}
%   \centering
%   \resizebox{1.\linewidth}{!}{
%   \begin{tabular}{l|c|c|c|c}
%          & \multicolumn{2}{c|}{TTM} & \multicolumn{2}{c}{MeMViT\cite{wu2022memvit}} \\
%         \toprule
%         Model & mAP & +GFLOPS & mAP & +GFLOPS \\
%         \midrule
%         % Pooling + sampling & 21.63 & 0.103 \\
%         Backbone & 25.2 & & 26.2 &    \\
%         + Memory & 27.9 (+2.7) & +0.8 & 28.5 (+2.3) & +1.3 \\
%         \bottomrule
%   \end{tabular}
%   }
%   \caption{Results on AVA 2.2, with the vanilla Kinetics-400 pretraining. We show FLOPS added by the sequential modeling per step. ViViT-B is the backbone for TTM, while MeMViT uses MViT.}
%   \label{table:ava}
% \end{table}

\begin{table}[h]
  \centering
  \small
  \begin{tabular}{l|c|c}
        \toprule
        Model & mAP & +GFLOPS \\
        \midrule
        % Pooling + sampling & 21.63 & 0.103 \\
        MViT & 26.2 & - \\
        ~~+ memory (i.e., MeMViT \cite{wu2022memvit}) & 28.5 (+2.3) & 1.3 \\
        \midrule
        ViViT-B & 25.2 & - \\
        ~~+ TTM per video & 27.9 (+2.7)  & 0.8 \\
        ~~+ TTM per box (\# layers=1) & 31.3 (+6.1)  & 1.0 \\
        ~~+ TTM per box (\# layers=4) & 31.5 (+6.3)  & 2.0 \\
        \bottomrule
  \end{tabular}
  \caption{Results on AVA 2.2, with the vanilla Kinetics-400 pretraining. We show FLOPS added by the sequential modeling per step. ViViT-B is the backbone for TTM, while MeMViT uses MViT.}
  \label{table:ava}
\end{table}

\begin{figure*}
    \centering
  \centering
  \includegraphics[width=.2\linewidth]{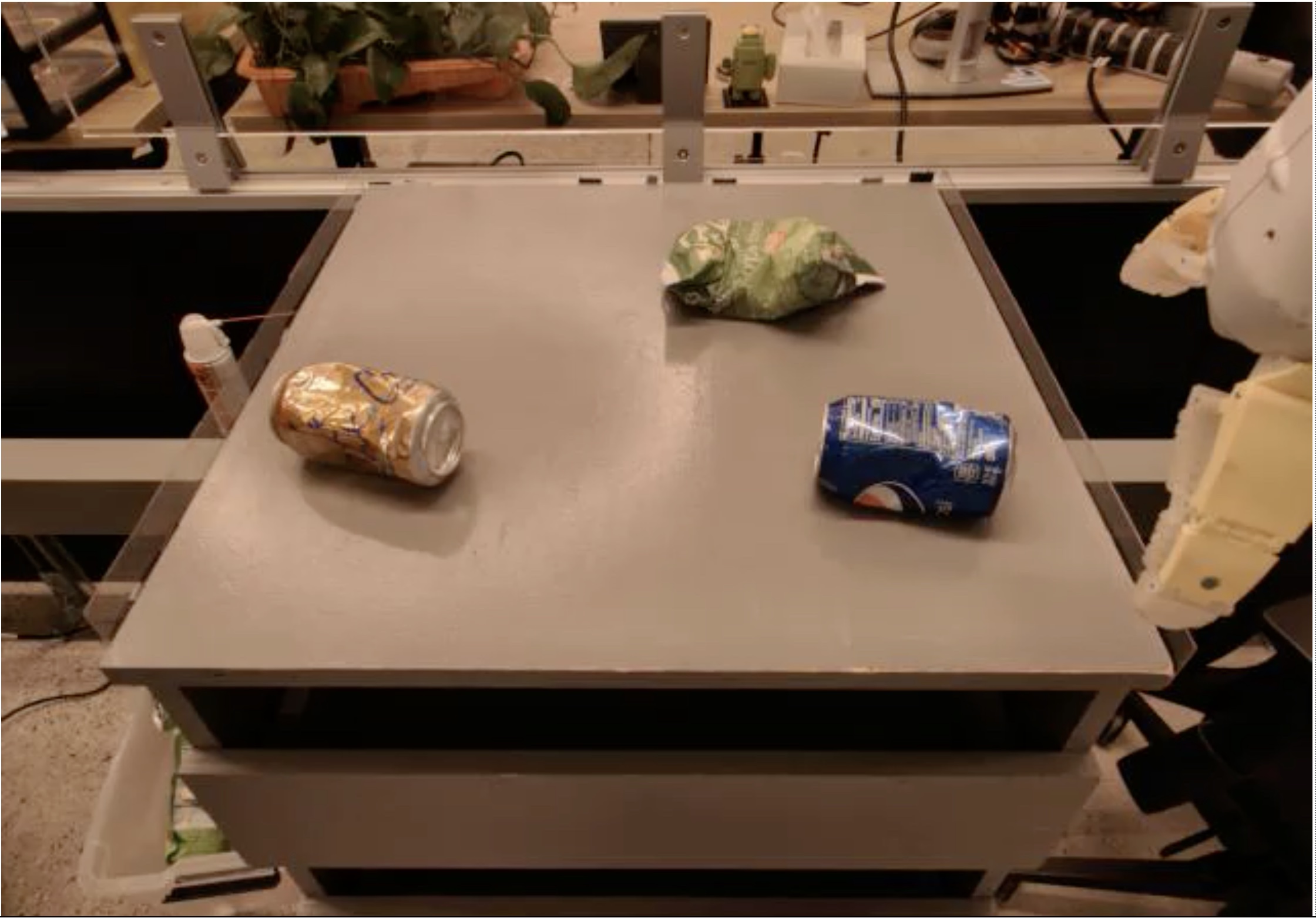}
  \includegraphics[width=.2\linewidth]{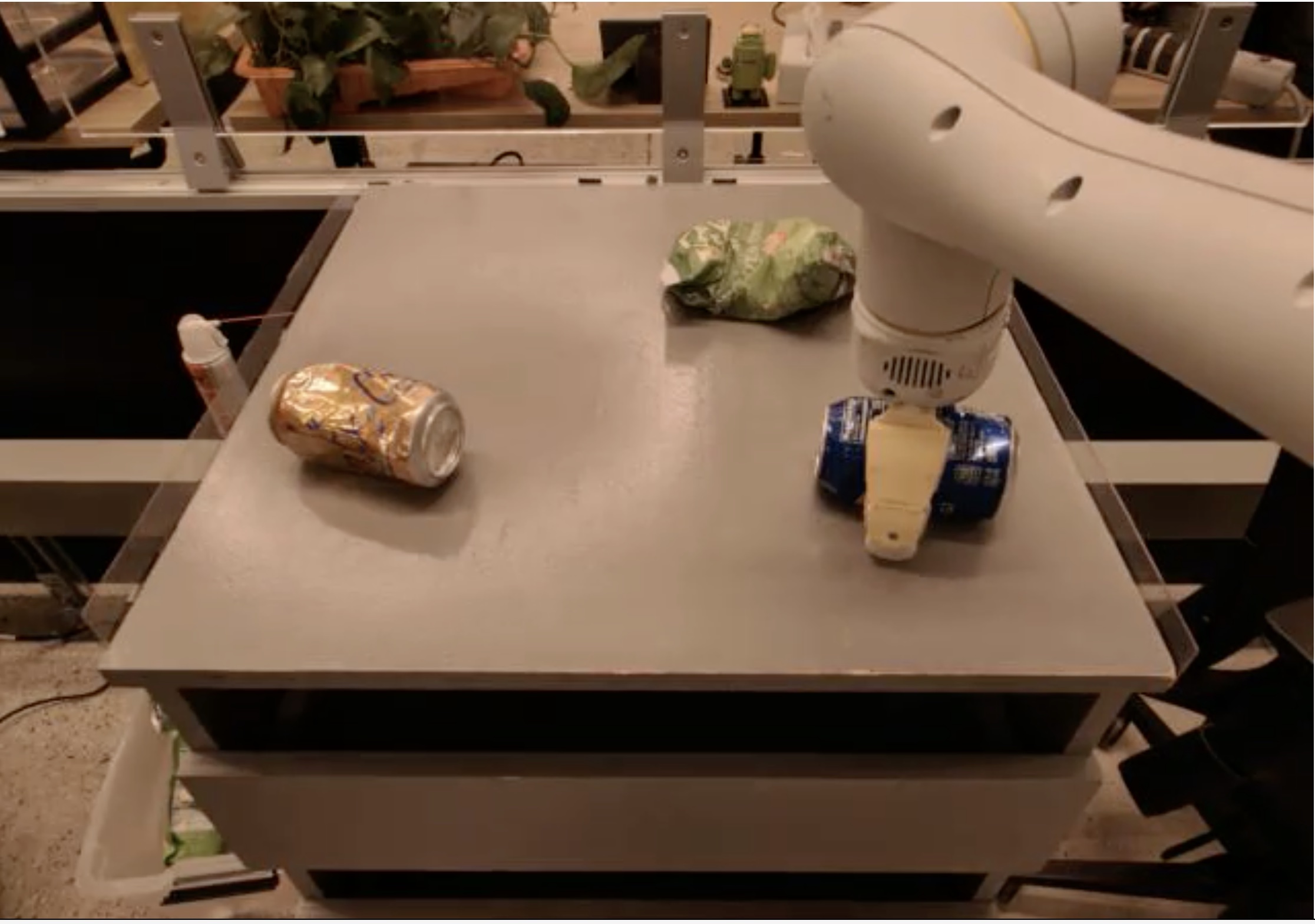}
  \includegraphics[width=.2\linewidth]{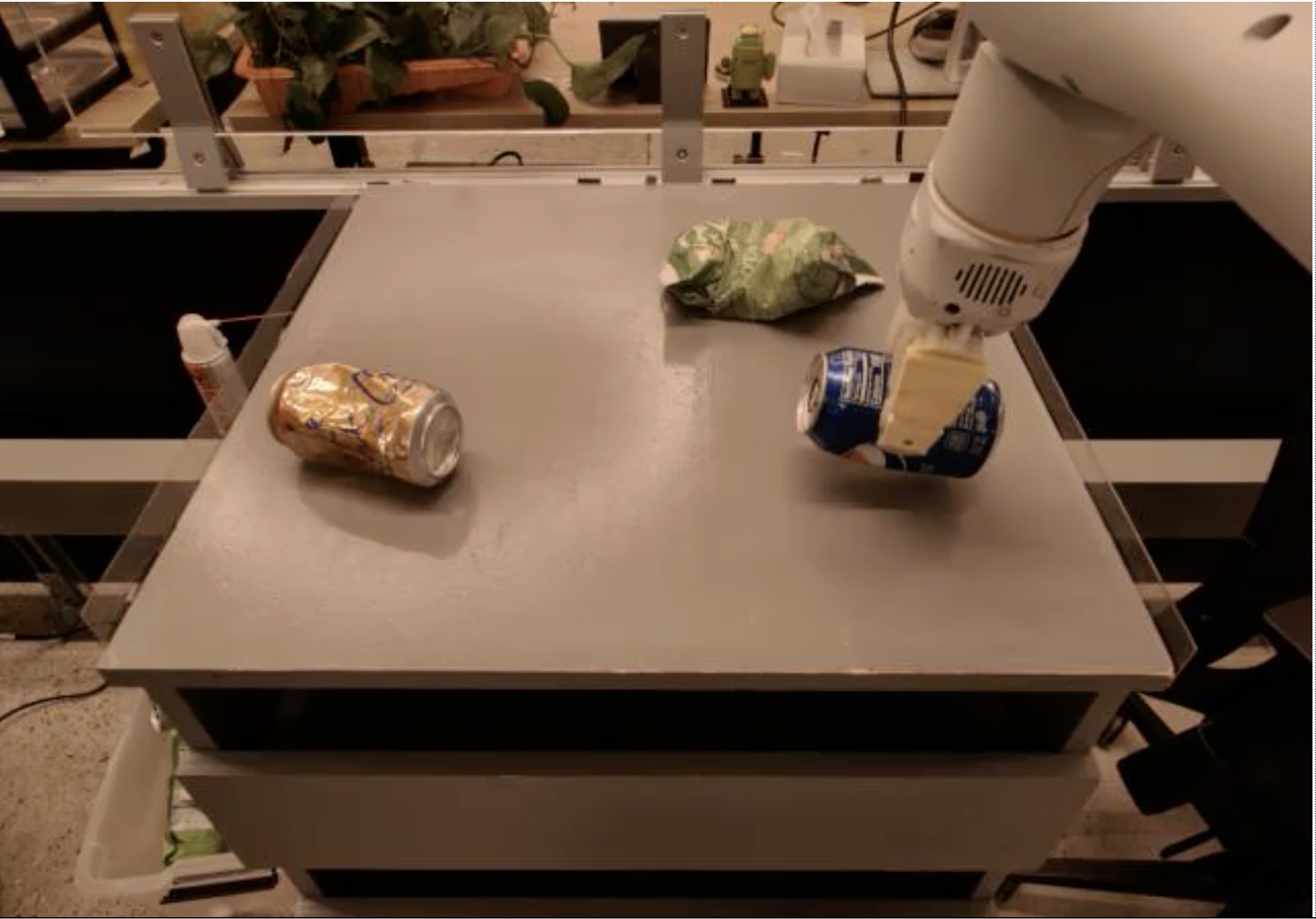}
  \includegraphics[width=.2\linewidth]{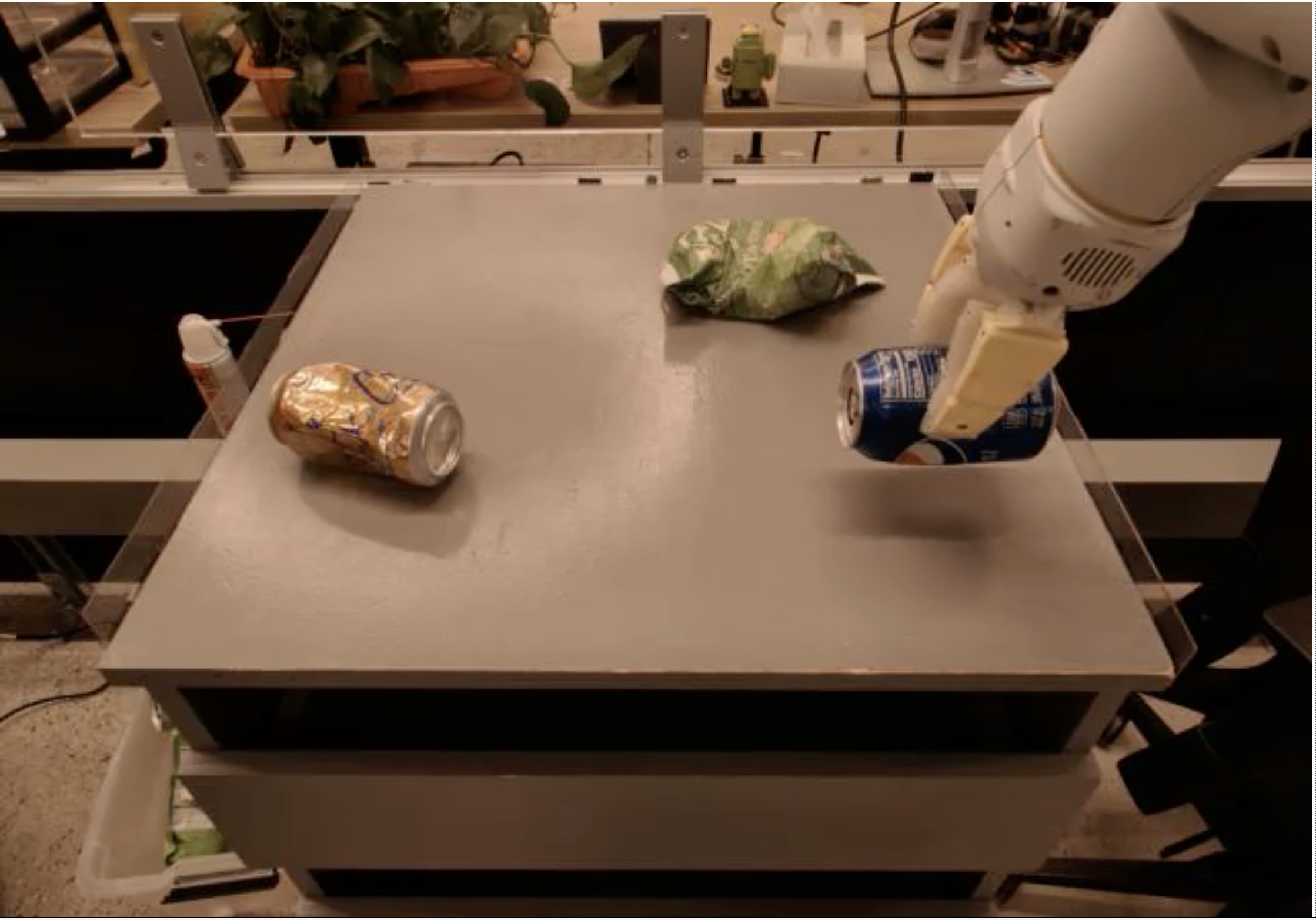}
  \includegraphics[width=.2\linewidth]{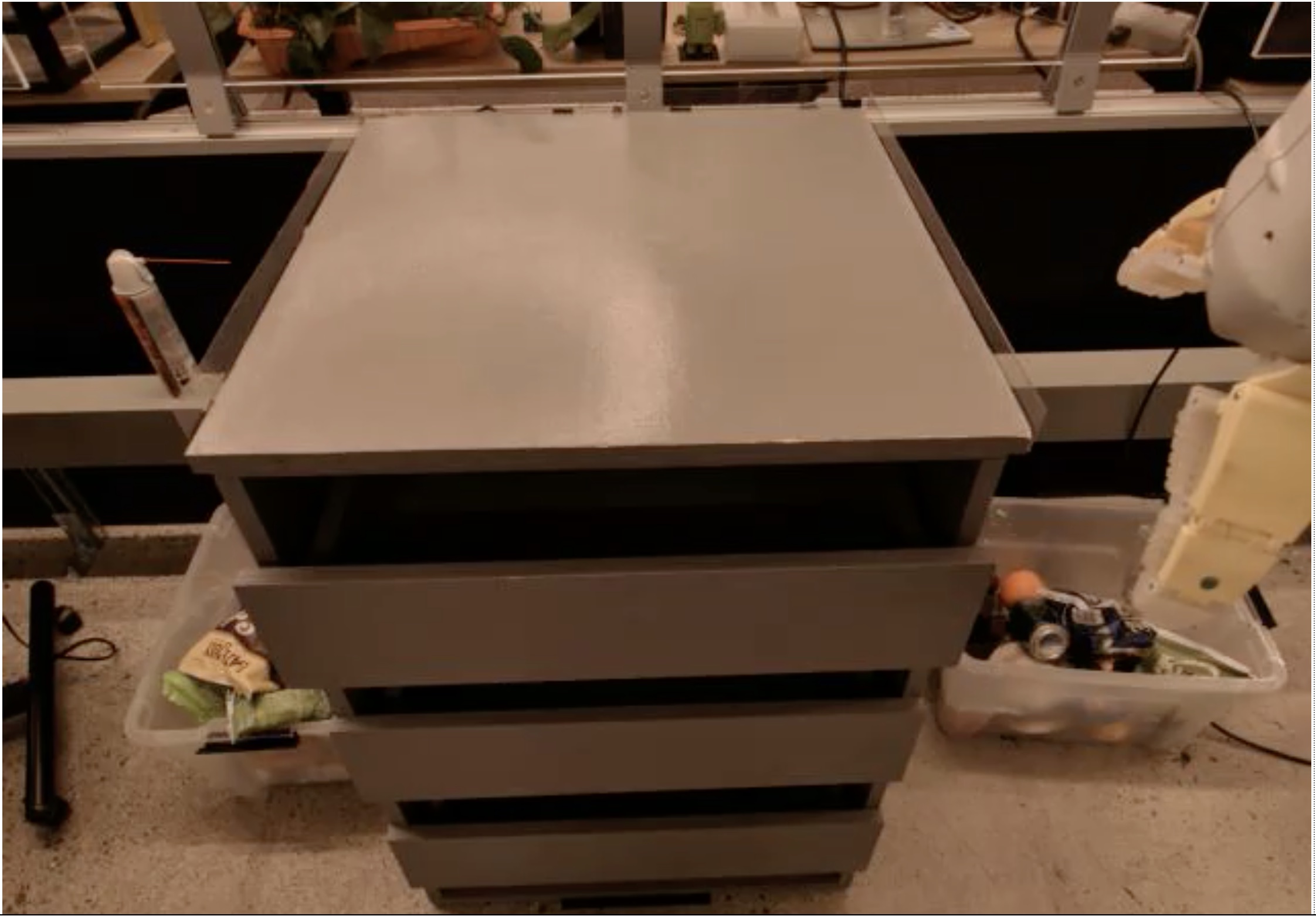}
  \includegraphics[width=.2\linewidth]{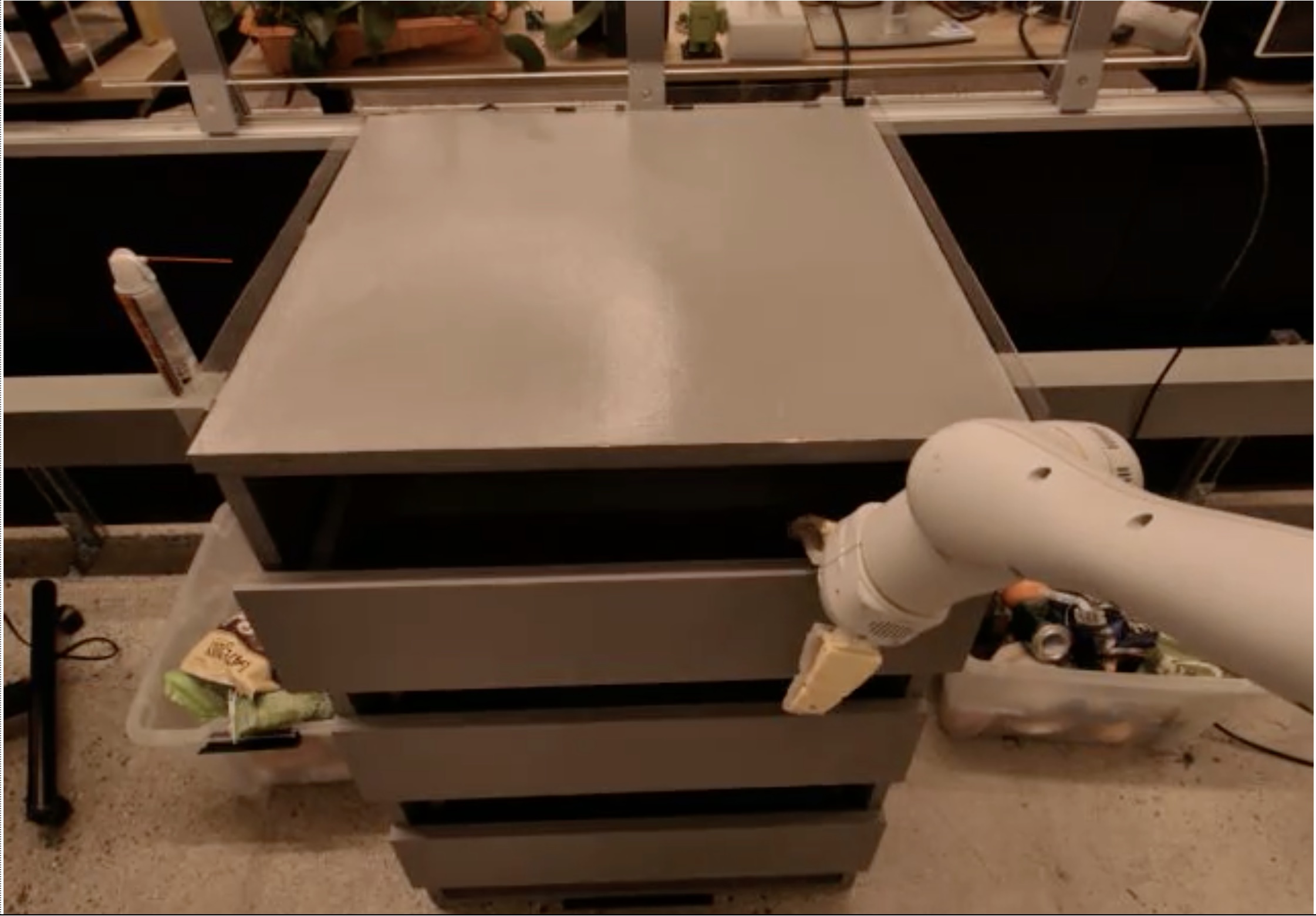}
  \includegraphics[width=.2\linewidth]{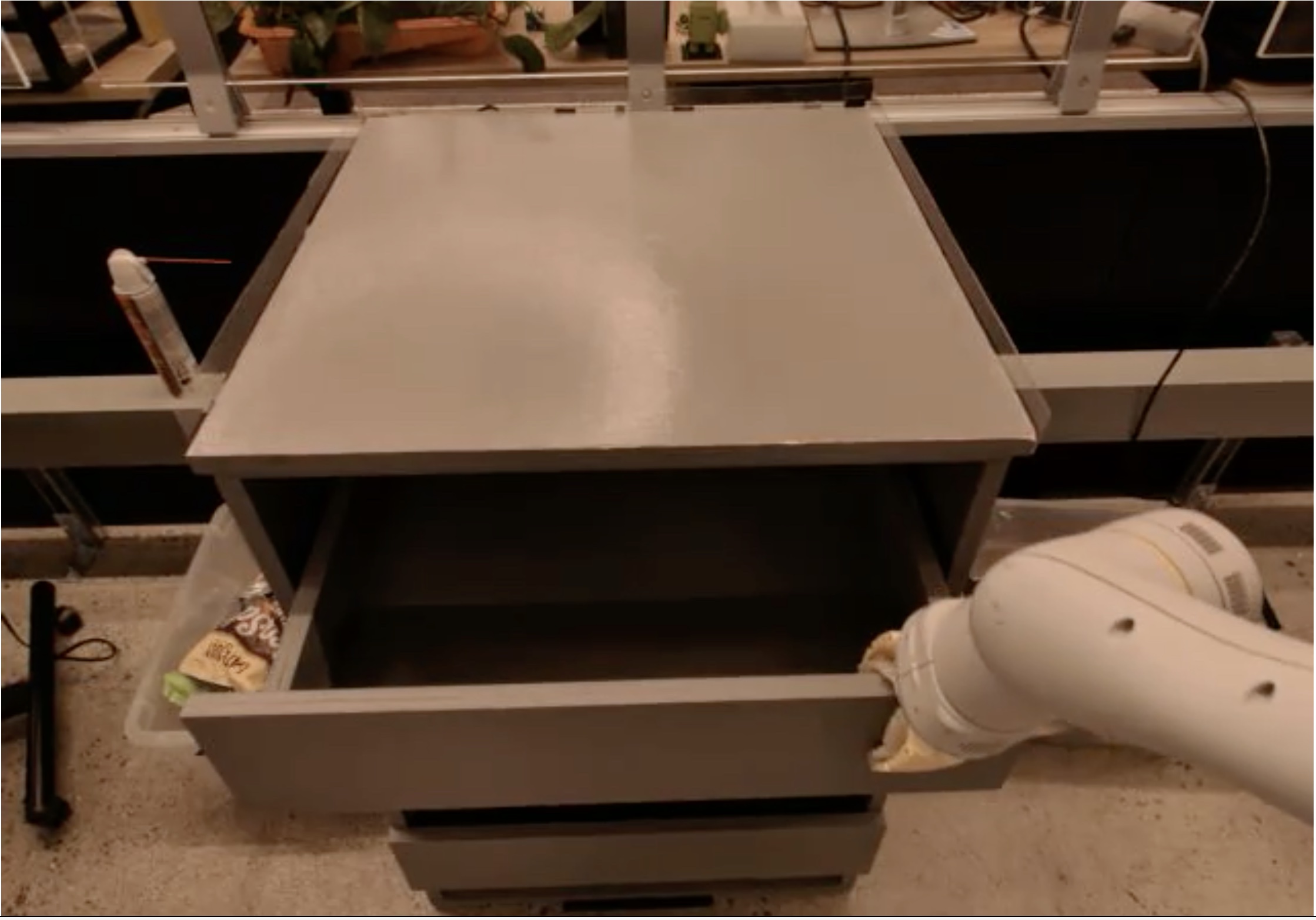}
  \includegraphics[width=.2\linewidth]{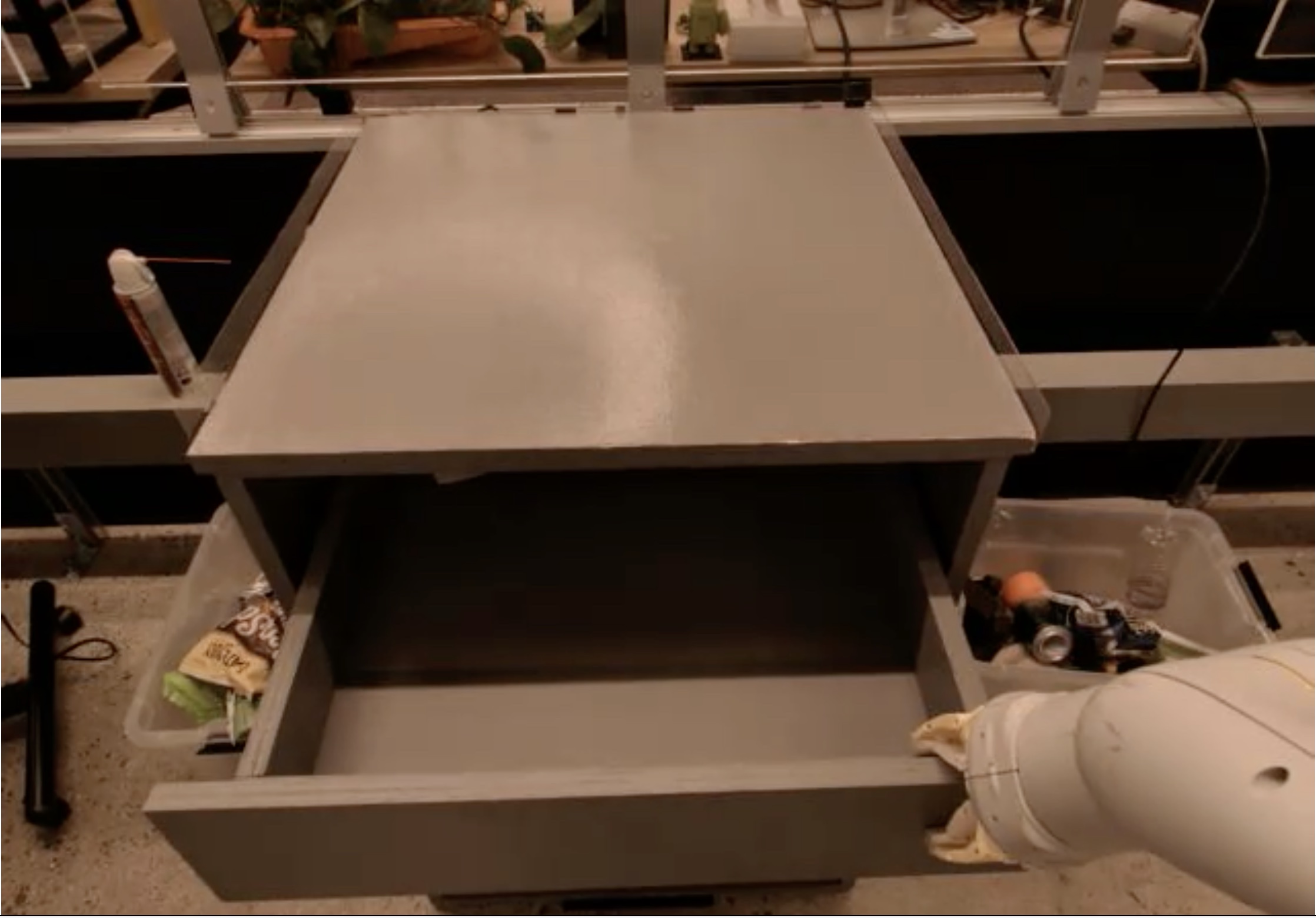}
    \caption{Example input frames from the robot task ``pick pepsi can'' (top) and ``open top drawer'' (bottom). We learn one action policy, which needs to cover all 551 task instructions.}
    \label{fig:robot-images1}
\end{figure*}

\subsubsection{Spatio-temporal Activity Detection Results}

We conducted an additional experiment on AVA \cite{gu2018ava} to confirm the benefit of TTM. We followed the lightest setting of pretraining the backbone with Kinetics-400. Identical to our Charades experiments, ViViT-B was used as the backbone. The only difference is that it was trained by providing video segments of 4 steps (i.e., $32\times4$ frames) at a time, due to the memory constraint. The bounding box proposals are obtained with the SlowFast network \cite{feichtenhofer2019slowfast}, and the TTM was responsible for computing the feature mask to be pooled per bounding box.

Table \ref{table:ava} compares the TTM with its backbone. We are able to observe the benefit of TTM, which improves the accuracy by managing the external memory, with a little added compute. We also provided its comparison against MeMViT \cite{wu2022memvit}, which also presents a Transformer-based memory mechanism for sequential visual data. Note that there are various ways to further boost the accuracy numbers orthogonal to the proposed sequential modeling. This includes the use of larger pretraining dataset \cite{wu2022memvit}, use of bigger backbones \cite{wu2022memvit, tong2022videomae}, and better pretraining with self-supervised losses \cite{tong2022videomae}. What we confirm in this experiment is the relative gain over the backbone; TTM with a small added compute, by utilizing the memory, improves AVA activity detection.

\begin{figure*}
\begin{minipage}{0.69\textwidth}
    \centering
    \includegraphics[width=0.9\linewidth]{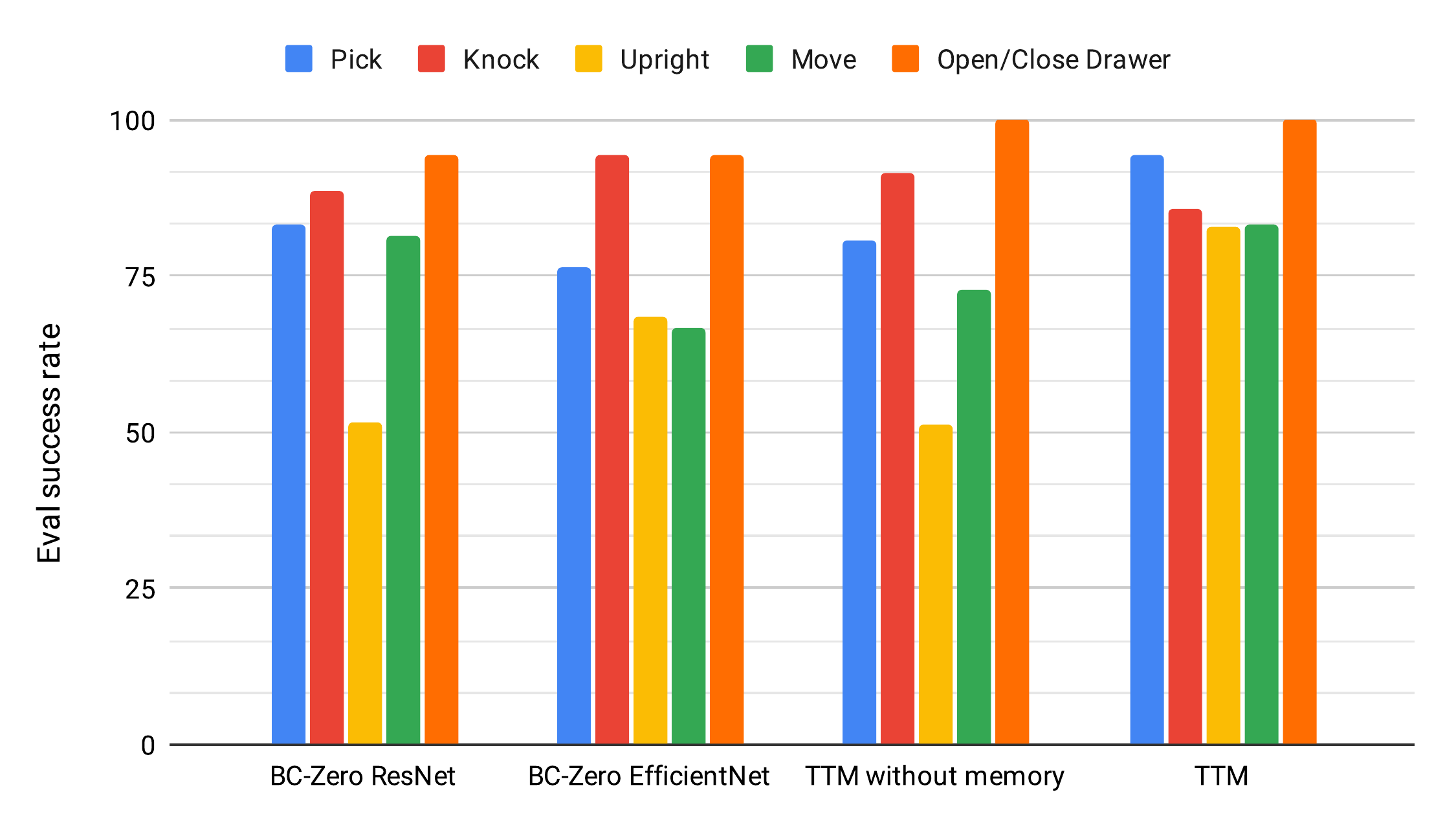}
    \caption{Real robot experiment; per-task success rates.}
    \label{fig:robot-results}
\end{minipage}
\begin{minipage}{0.3\textwidth}
  \centering
  \begin{tabular}{l|c}
        \toprule
        Model & Success \\
        \midrule
        BC ResNet & 79.80 \\
        BC EfficientNet & 80.08 \\
        No memory & 79.26 \\
        TTM & 89.26 \\
        \bottomrule
  \end{tabular}
  \caption{Average task success rate in real-robot evaluation.}
  \label{table:robot-results}
\end{minipage}
\vspace{-\baselineskip}
\end{figure*}

\subsection{Robot Learning}

To study how TTM scales to a real-world robotic control setting, we integrate it to a real kitchen environment described in SayCan~\cite{ahn2022saycan}.
An \href{https://everydayrobots.com/}{Everyday Robots} robot, a mobile manipulator with RGB observations, is placed in an office kitchen to interact with common objects using concurrent~\cite{xiao2020nonblocking} continuous closed-loop control from pixels. Here, at each time step, the inputs to the model are an image from the robot's mounted camera and the task instruction in natural language (Fig.~\ref{fig:robot-images1}). The expected output is an action vector for robot arm and base control. The policy was trained under a supervised behavioral cloning (BC) setting with human demonstrations.

\vspace{-2mm}
\subsubsection{Dataset and Settings}
\label{dataset}

We use the dataset and settings as described in SayCan~\cite{ahn2022saycan} with the additional inclusion of controls for base motion. We collect the dataset to train imitation learning policies: a real-world dataset of teleoperated human demonstrations of successful policy rollouts filtered by engineered success detectors.
Such real2real setup includes a training dataset of 89,000 teleoperated episodes collected in a mock kitchen across 551 tasks involving skills like picking, placing, and manipulating furniture. Each task instruction (with different objects) were given in the form of a text sentence, and the robot was asked to learn a single model for all such tasks. A policy trained on the dataset is evaluated again in the same kitchen in real-time. The tasks are grouped into 5 different types, Pick, Knock, Upright, Move, and Open/Close Drawer, and we report success rate of each type.
%The sim2sim testing set up includes a simulated dataset for training, consisting of 518,000 successful episodes obtained from executing a behavorial cloning policy trained on real data in simulation. Policies trained on this simulated dataset are then evaluated in sim. The simulated robot inputs in training and evaluation are mimiced to look like real world inputs by applying a CycleGAN ~\cite{rao2020rl} transformation. 

\vspace{-2mm}
\subsubsection{Baselines and Implementation}
\label{arch}
We follow the learning framework described in SayCan ~\cite{ahn2022saycan}. Here the first baseline we benchmark against is the ResNet based control policy network, called BC-Zero ResNet, developed in ~\cite{jang2021bcz} and used in SayCan. A second baseline we consider is BC-Zero with the image trunk swapped for a pretrained EfficientNet~\cite{tan2019efficientnet}, while still applying FiLM conditioning ~\cite{perez2020film} for language as described in~\cite{jang2021bcz} for ResNet. We call this the BC-Zero EfficientNet. EfficientNet is computationally more efficient and pretraining on ImageNet improves object understanding. 

Against these baselines we benchmark the proposed TTM architecture. TTM treats EfficientNet outputs as the input tokens. The memory size of TTM was $m = 96$, the number of reads was $r = 16$, and $n = 48$ after aggregation. A Transformer with 8 layers was used as the processing unit by default, and the total of 8 steps were considered at a time.
We also compare TTM against its memory-less version, which uses the same framework and the compute to the TTM. The only difference is that the memory has been zeroed out.
%As shown in Fig.~\ref{fig:robot-results} and Tab.~\ref{table:robot-results}, we observe significant improvements with TTM in the task success rate, since knowledge of previous history is critical for this task.

%All architectures are then trained in a behaviour cloning set up on the real2real dataset described in \ref{dataset}. 

% \subsection{Simulation - Bootstrap Learning?}
% \TED{I can go into more or less detail about the real2sim self-imitation learning setting, let me know and I can reorganize this as needed.}\MICHAEL{Let's add sim2sim here.}

\subsection{Real-robot Results}
We compare TTM against the baselines discussed above including BC-Zero used in SayCan. 
%published baselines such as SayCan ~\cite{ahn2022saycan} and variations of it described in \ref{arch}.
Inputs to the models are images from the robot's mounted headcam, previous actions executed in the episode and natural language instruction for the task. Outputs are action vectors to control the robot in real evaluation, as discussed in Section \ref{dataset}.

Fig.~\ref{fig:robot-results} and Table~\ref{table:robot-results} show the results.
We observe significant improvements with TTM in the task success rate, since knowledge of previous history is critical for this task.
Fig.~\ref{fig:robot-images1} shows frames from the real robot in operation.

\section{Conclusion}

We introduce Token Turing Machines for sequential decision making. Token Turing Machines could be viewed as a modernisation of Neural Turing Machines, with memory reads/writes designed in terms of token summarisations. It has good perks of modern Transformer-based models while also benefiting from having an external memory: constant compute regardless of the length of the history. Such capability is particularly important in many sequential decision making and online inference problems, such as robot action policy learning. We confirmed its power on real-world tasks with challenging visual inputs: Charades activity localization, and vision-based robot action policy learning.

\vspace{-5pt}
\paragraph{Discussions:}

The applicability of TTMs themselves is broad, as they are generic sequential models designed to digest a large number of tokens. Our intention with this paper particularly has been to focus on computer vision problems with sequential visual data. The problems that we chose (i.e., spatio-temporal human action localisation and robot policy) are challenging as they require extremely long sequences of tokens: For our experiments on action localisation, we have 3136 tokens per step, multiplied by 6 steps, which is a total of 18816 tokens. This sequence length is therefore significantly larger than those in other domains and comparable to Long Range Arena \cite{tay2020long} (1000 to 16000 tokens), posing comparable yet different challenges.

We also emphasize that TTM is the first to show its applicability to videos among the NTM-based approaches.

\appendix
\section{Appendix}
% \subsection{Reproducibility \& Ethics}
% In this paper, we proposed a sequential, autoregressive transformer model. We plan to open-source the code.
% The applications of our model are quite general (although we applied to human activity recognition and robotic control, many other applications exist), and it is not possible to be aware of all of its use-cases.
% Therefore, it is important to be cognizant that each application has its own merits and societal impacts depending on the specifics of the system, and how it is built and deployed.

\subsection{Charades Training}

For Charades~\cite{sigurdsson2016hollywood} training, we initialize ViViT~\cite{arnab2021vivit} backbones with pretrained weights (Base model: JFT~\cite{sun2017revisiting} $\rightarrow$ Kinetics-400~\cite{kay2017kinetics}, Large model: JFT $\rightarrow$ Kinetics-600~\cite{carreira2018short}) and initialize TTM-head with random weights. We finetune models with a batch size of 32 and Adam optimizer~\cite{kingma2014adam} with an initial learning rate of 1e-4 (with backbone learning rate further scaled by 0.1) and a cosine schedule for 100 epochs on 32 TPUv3 cores. To prevent overfitting, we use color/scale jitter, random augmentations~\cite{cubuk2020randaugment} and mixup~\cite{zhang2017mixup}. Our implementation is based on Jax~\cite{jax2018github} and the Scenic library~\cite{dehghani2021scenic}. We use sigmoid cross-entropy loss with a label smoothing of 0.1. Our inputs contain 6 temporal steps, each with 32 frames of 224$\times$224 resolution, and the loss is applied to the last step only. This allows better training for the TTM memory module.

\subsection{Robot Policy Training}

In our robot experiments, all architectures are trained in a behavioral cloning setting on the real2real dataset described in Section \ref{dataset}. For the training of the policy, we follow experimental setup of SayCan \cite{ahn2022saycan}.

\iffalse
    \subsubsection*{Author Contributions}
    If you'd like to, you may include  a section for author contributions as is done
    in many journals. This is optional and at the discretion of the authors.
    
    \subsubsection*{Acknowledgments}
    Use unnumbered third level headings for the acknowledgments. All
    acknowledgments, including those to funding agencies, go at the end of the paper.
\fi

\balance
\bibliographystyle{ieee_fullname}
\bibliography{egbib}

\begin{thebibliography}{10}\itemsep=-1pt

\bibitem{ahn2022saycan}
Michael Ahn, Anthony Brohan, Noah Brown, Yevgen Chebotar, Omar Cortes, Byron
  David, Chelsea Finn, Keerthana Gopalakrishnan, Karol Hausman, Alex Herzog,
  Daniel Ho, Jasmine Hsu, Julian Ibarz, Brian Ichter, Alex Irpan, Eric Jang,
  Rosario~Jauregui Ruano, Kyle Jeffrey, Sally Jesmonth, Nikhil Joshi, Ryan
  Julian, Dmitry Kalashnikov, Yuheng Kuang, Kuang-Huei Lee, Sergey Levine, Yao
  Lu, Linda Luu, Carolina Parada, Peter Pastor, Jornell Quiambao, Kanishka Rao,
  Jarek Rettinghouse, Diego Reyes, Pierre Sermanet, Nicolas Sievers, Clayton
  Tan, Alexander Toshev, Vincent Vanhoucke, Fei Xia, Ted Xiao, Peng Xu, Sichun
  Xu, and Mengyuan Yan.
\newblock Do as i can and not as i say: Grounding language in robotic
  affordances.
\newblock In {\em arXiv preprint arXiv:2204.01691}, 2022.

\bibitem{akbari2021vatt}
Hassan Akbari, Liangzhe Yuan, Rui Qian, Wei-Hong Chuang, Shih-Fu Chang, Yin
  Cui, and Boqing Gong.
\newblock Vatt: Transformers for multimodal self-supervised learning from raw
  video, audio and text.
\newblock {\em Advances in Neural Information Processing Systems},
  34:24206--24221, 2021.

\bibitem{arnab2021vivit}
Anurag Arnab, Mostafa Dehghani, Georg Heigold, Chen Sun, Mario Lu{\v{c}}i{\'c},
  and Cordelia Schmid.
\newblock Vivit: A video vision transformer.
\newblock In {\em ICCV}, 2021.

\bibitem{bertasius2021space}
Gedas Bertasius, Heng Wang, and Lorenzo Torresani.
\newblock Is space-time attention all you need for video understanding?
\newblock In {\em ICML}, volume~2, page~4, 2021.

\bibitem{borgeaud2022improving}
Sebastian Borgeaud, Arthur Mensch, Jordan Hoffmann, Trevor Cai, Eliza
  Rutherford, Katie Millican, George~Bm Van Den~Driessche, Jean-Baptiste
  Lespiau, Bogdan Damoc, Aidan Clark, et~al.
\newblock Improving language models by retrieving from trillions of tokens.
\newblock In {\em International Conference on Machine Learning}, 2022.

\bibitem{jax2018github}
James Bradbury, Roy Frostig, Peter Hawkins, Matthew~James Johnson, Chris Leary,
  Dougal Maclaurin, George Necula, Adam Paszke, Jake Vander{P}las, Skye
  Wanderman-{M}ilne, and Qiao Zhang.
\newblock {JAX}: composable transformations of {P}ython+{N}um{P}y programs,
  2018.

\bibitem{burtsev2020memory}
Mikhail~S Burtsev, Yuri Kuratov, Anton Peganov, and Grigory~V Sapunov.
\newblock Memory transformer.
\newblock {\em arXiv preprint arXiv:2006.11527}, 2020.

\bibitem{caba2015activitynet}
Fabian Caba~Heilbron, Victor Escorcia, Bernard Ghanem, and Juan Carlos~Niebles.
\newblock Activitynet: A large-scale video benchmark for human activity
  understanding.
\newblock In {\em Proceedings of the ieee conference on computer vision and
  pattern recognition}, pages 961--970, 2015.

\bibitem{carreira2018short}
Joao Carreira, Eric Noland, Andras Banki-Horvath, Chloe Hillier, and Andrew
  Zisserman.
\newblock A short note about kinetics-600, 2018.

\bibitem{carreira2017quo}
Joao Carreira and Andrew Zisserman.
\newblock Quo vadis, action recognition? a new model and the kinetics dataset.
\newblock In {\em proceedings of the IEEE Conference on Computer Vision and
  Pattern Recognition}, pages 6299--6308, 2017.

\bibitem{child2019generating}
Rewon Child, Scott Gray, Alec Radford, and Ilya Sutskever.
\newblock Generating long sequences with sparse transformers.
\newblock {\em arXiv preprint arXiv:1904.10509}, 2019.

\bibitem{choromanski2020rethinking}
Krzysztof Choromanski, Valerii Likhosherstov, David Dohan, Xingyou Song,
  Andreea Gane, Tamas Sarlos, Peter Hawkins, Jared Davis, Afroz Mohiuddin,
  Lukasz Kaiser, et~al.
\newblock Rethinking attention with performers.
\newblock {\em arXiv preprint arXiv:2009.14794}, 2020.

\bibitem{chung2014empirical}
Junyoung Chung, Caglar Gulcehre, KyungHyun Cho, and Yoshua Bengio.
\newblock Empirical evaluation of gated recurrent neural networks on sequence
  modeling.
\newblock {\em arXiv preprint arXiv:1412.3555}, 2014.

\bibitem{cordonnier2021differentiable}
Jean-Baptiste Cordonnier, Aravindh Mahendran, Alexey Dosovitskiy, Dirk
  Weissenborn, Jakob Uszkoreit, and Thomas Unterthiner.
\newblock Differentiable patch selection for image recognition.
\newblock In {\em CVPR}, 2021.

\bibitem{cubuk2020randaugment}
Ekin~D Cubuk, Barret Zoph, Jonathon Shlens, and Quoc~V Le.
\newblock Randaugment: Practical automated data augmentation with a reduced
  search space.
\newblock In {\em Proceedings of the IEEE/CVF conference on computer vision and
  pattern recognition workshops}, pages 702--703, 2020.

\bibitem{dai2021ctrn}
Rui Dai, Srijan Das, and Francois Bremond.
\newblock Ctrn: Class temporal relational network for action detection.
\newblock In {\em BMVC 2021-The British Machine Vision Conference}, 2021.

\bibitem{dai2022ms}
Rui Dai, Srijan Das, Kumara Kahatapitiya, Michael~S Ryoo, and Francois Bremond.
\newblock Ms-tct: Multi-scale temporal convtransformer for action detection.
\newblock In {\em Proceedings of the IEEE/CVF Conference on Computer Vision and
  Pattern Recognition}, pages 20041--20051, 2022.

\bibitem{dai2021pdan}
Rui Dai, Srijan Das, Luca Minciullo, Lorenzo Garattoni, Gianpiero Francesca,
  and Francois Bremond.
\newblock Pdan: Pyramid dilated attention network for action detection.
\newblock In {\em Proceedings of the IEEE/CVF Winter Conference on Applications
  of Computer Vision}, pages 2970--2979, 2021.

\bibitem{dai2019transformer}
Zihang Dai, Zhilin Yang, Yiming Yang, Jaime Carbonell, Quoc~V Le, and Ruslan
  Salakhutdinov.
\newblock Transformer-xl: Attentive language models beyond a fixed-length
  context.
\newblock In {\em ACL}, 2019.

\bibitem{dehghani2021scenic}
Mostafa Dehghani, Alexey Gritsenko, Anurag Arnab, Matthias Minderer, and Yi
  Tay.
\newblock Scenic: A jax library for computer vision research and beyond.
\newblock In {\em Proceedings of the IEEE/CVF Conference on Computer Vision and
  Pattern Recognition (CVPR)}, pages 21393--21398, 2022.

\bibitem{didolkar2022temporal}
Aniket Didolkar, Kshitij Gupta, Anirudh Goyal, Nitesh~B. Gundavarapu, Alex
  Lamb, Nan~Rosemary Ke, and Yoshua Bengio.
\newblock Temporal latent bottleneck: Synthesis of fast and slow processing
  mechanisms in sequence learning.
\newblock In {\em NeurIPS}, 2022.

\bibitem{donahue2015long}
Jeffrey Donahue, Lisa Anne~Hendricks, Sergio Guadarrama, Marcus Rohrbach,
  Subhashini Venugopalan, Kate Saenko, and Trevor Darrell.
\newblock Long-term recurrent convolutional networks for visual recognition and
  description.
\newblock In {\em CVPR}, June 2015.

\bibitem{dosovitskiy2020}
Alexey Dosovitskiy, Lucas Beyer, Alexander Kolesnikov, Dirk Weissenborn,
  Xiaohua Zhai, Thomas Unterthiner, Mostafa Dehghani, Matthias Minderer, Georg
  Heigold, Sylvain Gelly, Jakob Uszkoreit, and Neil Houlsby.
\newblock An image is worth 16x16 words: Transformers for image recognition at
  scale.
\newblock {\em arXiv preprint arXiv:2010.11929}, 2020.

\bibitem{fan2021multiscale}
Haoqi Fan, Bo Xiong, Karttikeya Mangalam, Yanghao Li, Zhicheng Yan, Jitendra
  Malik, and Christoph Feichtenhofer.
\newblock Multiscale vision transformers.
\newblock In {\em Proceedings of the IEEE/CVF International Conference on
  Computer Vision}, pages 6824--6835, 2021.

\bibitem{fayyaz2022adaptive}
Mohsen Fayyaz, Soroush~Abbasi Koohpayegani, Farnoush~Rezaei Jafari, Sunando
  Sengupta, Hamid Reza~Vaezi Joze, Eric Sommerlade, Hamed Pirsiavash, and
  Juergen Gall.
\newblock Adaptive token sampling for efficient vision transformers.
\newblock In {\em ECCV}, 2022.

\bibitem{feichtenhofer2020x3d}
Christoph Feichtenhofer.
\newblock X3{D}: Expanding architectures for efficient video recognition.
\newblock In {\em Proceedings of the IEEE/CVF Conference on Computer Vision and
  Pattern Recognition}, pages 203--213, 2020.

\bibitem{feichtenhofer2019slowfast}
Christoph Feichtenhofer, Haoqi Fan, Jitendra Malik, and Kaiming He.
\newblock Slowfast networks for video recognition.
\newblock In {\em Proceedings of the IEEE/CVF international conference on
  computer vision}, pages 6202--6211, 2019.

\bibitem{ghosh2020stacked}
Pallabi Ghosh, Yi Yao, Larry Davis, and Ajay Divakaran.
\newblock Stacked spatio-temporal graph convolutional networks for action
  segmentation.
\newblock In {\em The IEEE Winter Conference on Applications of Computer
  Vision}, pages 576--585, 2020.

\bibitem{goyal2022coordination}
Anirudh Goyal, Aniket Didolkar, Alex Lamb, Kartikeya Badola, Nan~Rosemary Ke,
  Nasim Rahaman, Jonathan Binas, Charles Blundell, Michael Mozer, and Yoshua
  Bengio.
\newblock Coordination among neural modules through a shared global workspace.
\newblock In {\em ICLR}, 2022.

\bibitem{graves2014neural}
Alex Graves, Greg Wayne, and Ivo Danihelka.
\newblock Neural {Turing} machines.
\newblock {\em arXiv preprint arXiv:1410.5401}, 2014.

\bibitem{graves2016hybrid}
Alex Graves, Greg Wayne, Malcolm Reynolds, Tim Harley, Ivo Danihelka, Agnieszka
  Grabska-Barwi{\'n}ska, Sergio~G{\'o}mez Colmenarejo, Edward Grefenstette,
  Tiago Ramalho, John Agapiou, et~al.
\newblock Hybrid computing using a neural network with dynamic external memory.
\newblock {\em Nature}, 538(7626):471--476, 2016.

\bibitem{gu2018ava}
Chunhui Gu, Chen Sun, David~A Ross, Carl Vondrick, Caroline Pantofaru, Yeqing
  Li, Sudheendra Vijayanarasimhan, George Toderici, Susanna Ricco, Rahul
  Sukthankar, et~al.
\newblock Ava: A video dataset of spatio-temporally localized atomic visual
  actions.
\newblock In {\em Proceedings of the IEEE Conference on Computer Vision and
  Pattern Recognition}, pages 6047--6056, 2018.

\bibitem{guu2020retrieval}
Kelvin Guu, Kenton Lee, Zora Tung, Panupong Pasupat, and Mingwei Chang.
\newblock Retrieval augmented language model pre-training.
\newblock In {\em International Conference on Machine Learning}, 2020.

\bibitem{hochreiter2001gradient}
Sepp Hochreiter, Yoshua Bengio, Paolo Frasconi, J{\"u}rgen Schmidhuber, et~al.
\newblock Gradient flow in recurrent nets: the difficulty of learning long-term
  dependencies, 2001.

\bibitem{hochreiter1997long}
Sepp Hochreiter and J{\"u}rgen Schmidhuber.
\newblock Long short-term memory.
\newblock {\em Neural computation}, 9(8):1735--1780, 1997.

\bibitem{hutchins2022block}
DeLesley Hutchins, Imanol Schlag, Yuhuai Wu, Ethan Dyer, and Behnam Neyshabur.
\newblock Block-recurrent transformers.
\newblock In {\em NeurIPS}, 2022.

\bibitem{jaegle2021perceiver}
Andrew Jaegle, Felix Gimeno, Andy Brock, Oriol Vinyals, Andrew Zisserman, and
  Joao Carreira.
\newblock Perceiver: General perception with iterative attention.
\newblock In {\em ICML}, 2021.

\bibitem{jang2021bcz}
Eric Jang, Alex Irpan, Mohi Khansari, Daniel Kappler, Frederik Ebert, Corey
  Lynch, Sergey Levine, and Chelsea Finn.
\newblock Bc-z: Zero-shot task generalization with robotic imitation learning.
\newblock 2022.

\bibitem{kahatapitiya2021self}
Kumara Kahatapitiya, Zhou Ren, Haoxiang Li, Zhenyu Wu, and Michael~S Ryoo.
\newblock Self-supervised pretraining with classification labels for temporal
  activity detection.
\newblock {\em arXiv preprint arXiv:2111.13675}, 2021.

\bibitem{kahatapitiya2021coarse}
Kumara Kahatapitiya and Michael~S Ryoo.
\newblock Coarse-fine networks for temporal activity detection in videos.
\newblock In {\em Proceedings of the IEEE/CVF Conference on Computer Vision and
  Pattern Recognition}, pages 8385--8394, 2021.

\bibitem{kaiser2017learning}
{\L}ukasz Kaiser, Ofir Nachum, Aurko Roy, and Samy Bengio.
\newblock Learning to remember rare events.
\newblock {\em arXiv preprint arXiv:1703.03129}, 2017.

\bibitem{kay2017kinetics}
Will Kay, Joao Carreira, Karen Simonyan, Brian Zhang, Chloe Hillier, Sudheendra
  Vijayanarasimhan, Fabio Viola, Tim Green, Trevor Back, Paul Natsev, Mustafa
  Suleyman, and Andrew Zisserman.
\newblock The kinetics human action video dataset, 2017.

\bibitem{khandelwal2019generalization}
Urvashi Khandelwal, Omer Levy, Dan Jurafsky, Luke Zettlemoyer, and Mike Lewis.
\newblock Generalization through memorization: Nearest neighbor language
  models.
\newblock {\em arXiv preprint arXiv:1911.00172}, 2019.

\bibitem{kingma2014adam}
Diederik~P Kingma and Jimmy Ba.
\newblock Adam: A method for stochastic optimization.
\newblock {\em arXiv preprint arXiv:1412.6980}, 2014.

\bibitem{le2019learning}
Hung Le, Truyen Tran, and Svetha Venkatesh.
\newblock Learning to remember more with less memorization.
\newblock {\em arXiv preprint arXiv:1901.01347}, 2019.

\bibitem{liu2021swin}
Ze Liu, Yutong Lin, Yue Cao, Han Hu, Yixuan Wei, Zheng Zhang, Stephen Lin, and
  Baining Guo.
\newblock Swin transformer: Hierarchical vision transformer using shifted
  windows.
\newblock In {\em Proceedings of the IEEE/CVF International Conference on
  Computer Vision}, pages 10012--10022, 2021.

\bibitem{liu2022video}
Ze Liu, Jia Ning, Yue Cao, Yixuan Wei, Zheng Zhang, Stephen Lin, and Han Hu.
\newblock Video swin transformer.
\newblock In {\em Proceedings of the IEEE/CVF Conference on Computer Vision and
  Pattern Recognition}, pages 3202--3211, 2022.

\bibitem{mavroudi2020representation}
Effrosyni Mavroudi, Benjam{\'\i}n~B{\'e}jar Haro, and Ren{\'e} Vidal.
\newblock Representation learning on visual-symbolic graphs for video
  understanding.
\newblock In {\em European Conference on Computer Vision}, pages 71--90.
  Springer, 2020.

\bibitem{nagrani2021attention}
Arsha Nagrani, Shan Yang, Anurag Arnab, Aren Jansen, Cordelia Schmid, and Chen
  Sun.
\newblock Attention bottlenecks for multimodal fusion.
\newblock {\em Advances in Neural Information Processing Systems},
  34:14200--14213, 2021.

\bibitem{peng2021random}
Hao Peng, Nikolaos Pappas, Dani Yogatama, Roy Schwartz, Noah~A Smith, and
  Lingpeng Kong.
\newblock Random feature attention.
\newblock {\em arXiv preprint arXiv:2103.02143}, 2021.

\bibitem{perez2020film}
Ethan Perez, Florian Strub, Harm de Vries, Vincent Dumoulin, and Aaron~C.
  Courville.
\newblock Film: Visual reasoning with a general conditioning layer.
\newblock {\em CoRR}, abs/1709.07871, 2017.

\bibitem{piergiovanni2018learning}
AJ Piergiovanni and Michael~S Ryoo.
\newblock Learning latent super-events to detect multiple activities in videos.
\newblock In {\em Proceedings of the IEEE Conference on Computer Vision and
  Pattern Recognition}, pages 5304--5313, 2018.

\bibitem{piergiovanni2019temporal}
AJ Piergiovanni and Michael~S. Ryoo.
\newblock Temporal gaussian mixture layer for videos.
\newblock In {\em International Conference on Machine learning}, pages
  5152--5161, 2019.

\bibitem{rae2019compressive}
Jack~W Rae, Anna Potapenko, Siddhant~M Jayakumar, and Timothy~P Lillicrap.
\newblock Compressive transformers for long-range sequence modelling.
\newblock {\em arXiv preprint arXiv:1911.05507}, 2019.

\bibitem{rao2021dynamicvit}
Yongming Rao, Wenliang Zhao, Benlin Liu, Jiwen Lu, Jie Zhou, and Cho-Jui Hsieh.
\newblock Dynamicvit: Efficient vision transformers with dynamic token
  sparsification.
\newblock In {\em NeurIPS}, 2021.

\bibitem{ryoo2021tokenlearner}
Michael Ryoo, AJ Piergiovanni, Anurag Arnab, Mostafa Dehghani, and Anelia
  Angelova.
\newblock Tokenlearner: Adaptive space-time tokenization for videos.
\newblock In {\em NeurIPS}, 2021.

\bibitem{sigurdsson2016hollywood}
Gunnar~A Sigurdsson, G{\"u}l Varol, Xiaolong Wang, Ali Farhadi, Ivan Laptev,
  and Abhinav Gupta.
\newblock Hollywood in homes: Crowdsourcing data collection for activity
  understanding.
\newblock In {\em European Conference on Computer Vision}, pages 510--526.
  Springer, 2016.

\bibitem{sun2017revisiting}
Chen Sun, Abhinav Shrivastava, Saurabh Singh, and Abhinav Gupta.
\newblock Revisiting unreasonable effectiveness of data in deep learning era.
\newblock In {\em Proceedings of the IEEE international conference on computer
  vision}, pages 843--852, 2017.

\bibitem{tan2019efficientnet}
Mingxing Tan and Quoc Le.
\newblock Efficientnet: Rethinking model scaling for convolutional neural
  networks.
\newblock In {\em International conference on machine learning}, pages
  6105--6114. PMLR, 2019.

\bibitem{tay2020long}
Yi Tay, Mostafa Dehghani, Samira Abnar, Yikang Shen, Dara Bahri, Philip Pham,
  Jinfeng Rao, Liu Yang, Sebastian Ruder, and Donald Metzler.
\newblock Long range arena: A benchmark for efficient transformers.
\newblock {\em arXiv preprint arXiv:2011.04006}, 2020.

\bibitem{tay2020efficient}
Yi Tay, Mostafa Dehghani, Dara Bahri, and Donald Metzler.
\newblock Efficient transformers: A survey.
\newblock {\em ACM Computing Surveys (CSUR)}, 2022.

\bibitem{mlpmixer2021}
Ilya Tolstikhin, Neil Houlsby, Alexander Kolesnikov, Lucas Beyer, Xiaohua Zhai,
  Thomas Unterthiner, Jessica Yung, Andreas Steiner, Daniel Keysers, Jakob
  Uszkoreit, Mario Lucic, and Alexey Dosovitskiy.
\newblock Mlp-mixer: An all-mlp architecture for vision.
\newblock In {\em NeurIPS}, 2021.

\bibitem{tong2022videomae}
Zhan Tong, Yibing Song, Jue Wang, and Limin Wang.
\newblock Videomae: Masked autoencoders are data-efficient learners for
  self-supervised video pre-training.
\newblock {\em arXiv preprint arXiv:2203.12602}, 2022.

\bibitem{vaswani2017attention}
Ashish Vaswani, Noam Shazeer, Niki Parmar, Jakob Uszkoreit, Llion Jones,
  Aidan~N Gomez, {\L}ukasz Kaiser, and Illia Polosukhin.
\newblock Attention is all you need.
\newblock In {\em NeurIPS}, 2017.

\bibitem{wang2020linformer}
Sinong Wang, Belinda~Z Li, Madian Khabsa, Han Fang, and Hao Ma.
\newblock Linformer: Self-attention with linear complexity.
\newblock {\em arXiv preprint arXiv:2006.04768}, 2020.

\bibitem{wu2022memvit}
Chao-Yuan Wu, Yanghao Li, Karttikeya Mangalam, Haoqi Fan, Bo Xiong, Jitendra
  Malik, and Christoph Feichtenhofer.
\newblock Memvit: Memory-augmented multiscale vision transformer for efficient
  long-term video recognition.
\newblock In {\em CVPR}, 2022.

\bibitem{wu2022memorizing}
Yuhuai Wu, Markus~N Rabe, DeLesley Hutchins, and Christian Szegedy.
\newblock Memorizing transformers.
\newblock In {\em ICLR}, 2022.

\bibitem{xiao2020nonblocking}
Ted Xiao, Eric Jang, Dmitry Kalashnikov, Sergey Levine, Julian Ibarz, Karol
  Hausman, and Alexander Herzog.
\newblock Thinking while moving: Deep reinforcement learning with concurrent
  control.
\newblock {\em arXiv preprint arXiv:2004.06089}, 2020.

\bibitem{yeung2018every}
Serena Yeung, Olga Russakovsky, Ning Jin, Mykhaylo Andriluka, Greg Mori, and Li
  Fei-Fei.
\newblock Every moment counts: Dense detailed labeling of actions in complex
  videos.
\newblock {\em International Journal of Computer Vision}, 126(2-4):375--389,
  2018.

\bibitem{yin2022vit}
Hongxu Yin, Arash Vahdat, Jose~M Alvarez, Arun Mallya, Jan Kautz, and Pavlo
  Molchanov.
\newblock A-vit: Adaptive tokens for efficient vision transformer.
\newblock In {\em CVPR}, 2022.

\bibitem{zaheer2020big}
Manzil Zaheer, Guru Guruganesh, Kumar~Avinava Dubey, Joshua Ainslie, Chris
  Alberti, Santiago Ontanon, Philip Pham, Anirudh Ravula, Qifan Wang, Li Yang,
  et~al.
\newblock Big bird: Transformers for longer sequences.
\newblock {\em Advances in Neural Information Processing Systems},
  33:17283--17297, 2020.

\bibitem{zhang2017mixup}
Hongyi Zhang, Moustapha Cisse, Yann~N Dauphin, and David Lopez-Paz.
\newblock mixup: Beyond empirical risk minimization.
\newblock {\em arXiv preprint arXiv:1710.09412}, 2017.

\end{thebibliography}

%\newpage
%\input{text/appendix.tex}

\end{document}